%% file: main.tex
\documentclass[twoside]{article}

\usepackage{PRIMEarxiv}

\usepackage[round]{natbib}
\usepackage[utf8]{inputenc} 
\usepackage[T1]{fontenc}    
\usepackage{hyperref}       
\usepackage{url}            
\usepackage{booktabs}       
\usepackage{amsfonts}       
\usepackage{nicefrac}       
\usepackage{microtype}      
\usepackage{xcolor}         
\usepackage{algorithm}
\usepackage{algpseudocode}

\usepackage{amsmath}
\usepackage{amssymb}
\usepackage{mathtools}
\usepackage{amsthm}

\usepackage{microtype}
\usepackage{graphicx}
\usepackage{subfigure}
\usepackage{booktabs} 
\usepackage{tikz}
 \usetikzlibrary {arrows.meta}
\usepackage{multirow}
\usepackage{hhline}
\usepackage{siunitx}

\hypersetup{
    colorlinks=true,
    linkcolor=blue,
    filecolor=magenta,
    citecolor = blue,
    urlcolor=cyan,
    pdftitle={DEA},
    pdfpagemode=FullScreen,
}

\theoremstyle{plain}
\newtheorem{theorem}{Theorem}[section]
\newtheorem{proposition}[theorem]{Proposition}

\theoremstyle{definition}
\newtheorem{definition}[theorem]{Definition}

\theoremstyle{remark}

\newtheorem{example}[theorem]{Example}

\newcommand{\Etrain}{\operatorname{\mathbb{E}_{\text{train}}}}

\newcommand{\Enu}{\operatorname{\mathbb{E}_\nu}}
\newcommand{\Pnu}{\operatorname{\mathbb{P}_\nu}}

\newcommand{\argmin}{\operatorname{\text{argmin}}}

\newcommand{\Bmax}{\operatorname{\mathbf{B}^{\diamond}}}

\newcommand{\cmark}{\ding{51}}%
\newcommand{\xmark}{\ding{55}}%
\usepackage{pifont}

\title{Out-of-distribution robustness for multivariate analysis via causal regularisation}

\author{
  Homer Durand, Gherardo Varando, Nathan Mankovich, Gustau Camps-Valls \\
  Image and Processing Lab \\
  Universitat de Valencia \\
  Valencia\\
  \texttt{homer.durand@uv.es} \\
  \texttt{gherardo.varando@uv.es} \\
  \texttt{gustau.camps@uv.es}
}

\begin{document}
\maketitle

\begin{abstract}
We propose a regularisation strategy of classical machine learning algorithms rooted in causality that ensures robustness against distribution shifts. Building upon the anchor regression framework, we demonstrate how incorporating a straightforward regularisation term into the loss function of classical multivariate analysis algorithms, such as (orthonormalized) partial least squares, reduced-rank regression, and multiple linear regression, enables out-of-distribution generalisation. Our framework allows users to efficiently verify the compatibility of a loss function with the regularisation strategy. Estimators for selected algorithms are provided, showcasing consistency and efficacy in synthetic and real-world climate science problems. The empirical validation highlights the versatility of anchor regularisation, emphasizing its compatibility with multivariate analysis approaches and its role in enhancing replicability while guarding against distribution shifts. The extended anchor framework advances causal inference methodologies, addressing the need for reliable out-of-distribution generalisation. 
\end{abstract}

\section{INTRODUCTION}
Data sources in contemporary machine learning applications are often heterogeneous, leading to potential distribution shifts 
\citep{sugiyama2012machine,Shen2021}. This is a particularly relevant problem in computer vision \citep{Csurka2017}, healthcare \citep{Zhang2021}, Earth and climate sciences \citep{Tuia2016, Kellenberger2021}, and social sciences \citep{jin2023diagnosing}, as variations in data patterns can significantly impact model performance and generalisation in the  
out-of-distribution (OOD) setting, also referred to as domain generalisation 
\citep{Shen2021, Zhou2023}. 

In this work, we address the problem of predicting a target $Y\in \mathcal{Y}$ from covariates $X \in \mathcal{X}$ where the training data consists of samples from a subset $\mathcal{Q}$ of a broader class of distributions $\mathbb{Q}$. Our objective is to achieve strong performance across the entire class $\mathbb{Q}$. This challenge can be framed as the following minimax problem:
\begin{align}\label{Eq:Distrib_Robust_def}
    \argmin_{\mathbf{\Theta}} \sup_{\mathcal{Q} \in \mathbb{Q}}\mathbb{E}_{(X, Y) \sim \mathcal{Q}}[\mathcal{L}(X, Y; \mathbf{\Theta})],
\end{align}
where $\mathcal{L}(X, Y; \mathbf{\Theta})$ is a loss function over $X$ and $Y$ with parameters $\mathbf{\Theta}$.
The choice of the class $\mathbb{Q}$ is particularly important as different classes of distributions lead to various types of robustness.  The class $\mathbb{Q}$, often referred to as the uncertainty set, can be constrained using distributional metrics such as $f$-divergence \citep[see, e.g.,][]{Namkoong2016} or the Wasserstein distance \citep[see, e.g.,][]{Esfahani2015}. The challenge of OOD 
generalisation is also closely related to the field of transfer learning \citep[see, e.g.,][]{Zhuang2021}. For an in-depth review of OOD generalisation, we recommend consulting \citep[][]{Shen2021, Zhou2023}. 

Of particular interest for this work, causal inference can be understood as a distributionally robust estimation where the class $\mathbb{Q}$ is the class of all distributions arising from arbitrarily large interventions on the variables of an SCM \citep[see][]{Meinshausen2018, Shen2021}. Notably, the Instrumental Variable (IV) regression exhibits  robustness to arbitrarily strong interventions on the instruments under some structural assumptions \citep{bowden1990instrumental}.
However, it might be overconservative to pursue algorithms robust to arbitrarily strong interventions, especially when assuming access to knowledge regarding the intervention strength. 
\cite{Rothenhäusler2018} address this challenge by constraining the intervention to exogenous (so-called \emph{anchor}) variables and by bounding the intervention strength. This approach, known as Anchor Regression (AR), results in a straightforward regularisation of the Ordinary Least Squares (OLS) algorithm, enabling robust linear regression.

More concretely, in its original form, AR can be written, in the population case, as
\begin{align*}
    b^\gamma = \argmin_\gamma & \mathbb{E}\left[((I - P_A)(Y - X^T b))^2\right] \\
    &+ \gamma \mathbb{E}\left[(P_A(Y - X^T b))^2\right],
\end{align*}
with $Y \in \mathbb{R}$. The second term acts as a regulariser that, for $\gamma > 1$, \textit{enforces the minimization of the residuals' projection onto the linear space defined by $A$}. This term is akin to the two-stage least squares used in IV regression, as it promotes decorrelation between the anchor and the residuals, thereby making the model more robust to distribution shifts arising from interventions on the anchor. This regularisation approach has proven useful in various applications, notably in environmental fields where \cite{Oberst2021} applied it to air-quality prediction, and \cite{Sippel2021} and \cite{Szekely2022} utilized it for the Detection and Attribution (D\&A) of climate change.

Several AR extensions have been proposed, demonstrating the theoretical versatility of this framework. 
Notable examples include Kernel Anchor Regression  \citep{shi2023}, which extends AR to nonlinear SCMs by working on reproducing kernel Hilbert spaces; targeted anchor regression~\citep{Oberst2021}, which exploits prior knowledge of the direction of intervention on the anchor variables; and proxy anchor regression~\citep{Oberst2021}, which addresses cases where the anchor is unobserved but proxy variables are available. 
In \citet{Kook2021}, authors introduce a distributional extension of AR, expanding beyond the $\ell_2$ loss to accommodate censored or discrete targets.

Previous literature has focused on regression problems with one-dimensional target variables and ordinary least squares as the loss function. Building upon the original approach of AR, we introduce a regularisation strategy applicable to a broader class of algorithms, including some classical MultiVariate Analysis (MVA) techniques \citep{bilodeau1999theory, Arenas-Garcia2013, Borchani2015}, also known as multi-output algorithms. This contributes on three fronts. First, we extend the applicability of AR algorithms to various loss functions, thereby ensuring compatibility and robustness against distribution shifts induced by constrained interventions on anchor variables. Second, we redefine conventional MVA algorithms within the anchor framework, encompassing techniques such as Partial Least Squares (PLS), Orthonormalized PLS (OPLS), Reduced Rank Regression (RRR), and Multilinear Regression (MLR). This adaptation ensures that their respective loss functions are \emph{anchor-compatible}, and the application of straightforward regularisation provides robustness during testing. This, allows us to leverage the power and versatility of a diverse array of MVA algorithms while providing robustness guarantees against distributional shifts. Third, by introducing a new class of \textit{anchor-compatible} loss functions, we offer a simple and straightforward method to determine whether an algorithm could benefit from anchor regularization and its generalization properties.

The code used in this work can be found on \href{https://github.com/homerdurand/anchorMVA}{GitHub}. 

\section{ANCHOR FRAMEWORK}\label{sec:AR}

For notation, bold capital letters $\mathbf{U}$ represent matrices, bold lowercase letters $\mathbf{u}$ denote vectors, regular lowercase letters $u$ signify scalars, and uppercase letters $U$ represent random variables (potentially multivariate) which we assume are centered. Given two random vector $U$ and $V$ we define $UV$ as the row-wise stacking operation $UV = (U^T, V^T)^T$. The symbol $C_{U V}$ denotes Kronecker products involving random variables, i.e. $C_{U V} =UV \otimes UV$
and to simplify the notation we write $C_{U} = U \otimes U$. Since we assume centred random variables $\Sigma_{UV} = \mathbb{E}[C_{UV}]$ and $\Sigma_U = \mathbb{E}[C_U]$. By a slight abuse of notation, we denote $C_{UV|A} = \mathbb{E}\left[UV|A \right] \otimes  \mathbb{E}\left[UV|A\right]$ and $\Sigma_{UV|A}$ its expectation.\footnote{Not to be confused with the conditional covariance matrix $\mathrm{Cov}(UV|A)$. Our notation satisfies the total variance-covariance formula $\Sigma_{UV}= \mathbb{E}[\mathrm{Cov}(UV|A)] + \Sigma_{UV|A}$.}
Again for simplification, we write $\sigma_{U} = \mathbb{E}[C_U]$ and $\sigma_{U V} = \mathbb{E}[U \otimes V]$.
We denote the identity matrix with $\mathbf{I}$, and its dimension is implied by the context when not explicitly stated. Finally, $\|.\|_F$ represents the Froebinius norm.  

\subsection{The Class of Anchor SCM}\label{Subsecec:class_anchor_SCM}

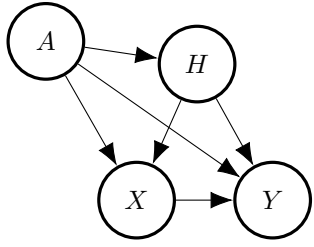
\begin{figure}
    \centering
    \begin{tikzpicture}[auto, scale=12,
     node distance = 18mm and 6mm,
    normaledge/.style ={arrows=-{Latex[length=3mm]}},
    dashededge/.style ={dashed, arrows=-{Latex[length=3mm]}}, 
    node/.style={circle,inner sep=1mm,minimum size=1cm,draw,black,very thick,text=black}]

     \node [node] (x) {$X$};
     \node [node] (y) [right of = x] {$Y$};
     \node [node] (h) [above of = x, xshift=8mm]  {$H$};
     \node [node] (a) [above of = x, xshift=-12mm, yshift=3mm]  {$A$};

     \path[normaledge] (a) edge (x);
     \path[normaledge] (a) edge (y);
     \path[normaledge] (a) edge (h);
     \path[normaledge] (h) edge (x);
     \path[normaledge] (h) edge (y);
     \path[normaledge] (x) edge (y);

     
    \end{tikzpicture}
    \caption{Directed Acyclic Graph (DAG) analyzed in this work, as induced by the Structural Causal Model (SCM) described in~\eqref{Eq:anchor-model}. The directions of the arrows between $X$, $Y$, and $H$ are flexible, provided that the graph remains acyclic. All possible configurations of the DAG can be seen in Fig.~\ref{fig:dag_variations} in Appendix.
    }

    \label{fig:dag_scm}
\end{figure}

In the following, we assume the Directed Acyclic Graph (DAG) in Fig.~\ref{fig:dag_scm}, where $X \in \mathbb{R}^d$ denotes the observed covariates, $Y \in \mathbb{R}^p$ represents the response (or target) variables, $H \in \mathbb{R}^r$ are unobserved variables potentially confounding the causal relation between $X$ and $Y$ and $A\in \mathbb{R}^{q}$ the exogenous (so-called \emph{anchor}) variables. We assume that the distribution of $(X, Y, H)$ is entailed by the following linear SCM $\mathcal{C}$:
\begin{align}\label{Eq:anchor-model}
    \begin{pmatrix} X \\ Y \\ H \end{pmatrix} = \mathbf{B} \begin{pmatrix}X \\Y\\H \end{pmatrix} + \varepsilon + \mathbf{M} A,
\end{align}

where $\mathbf{B} \in \mathbb{R}^{(d+p+r) \times (d+p+r)}$ and $\mathbf{M} \in \mathbb{R}^{(d+p+r)\times q}$ are unknown constant matrices and $\varepsilon \in \mathbb{R}^{d + p + r}$ is a vector of random noise. We assume that $A$ and $\varepsilon$ are independent and have finite variances, $\varepsilon$ components are independent, and $X$ and $Y$ have zero mean. 
The model in Eq.~\ref{Eq:anchor-model} generalises the Instrumental Variables setting, offering a flexible framework that encompasses any linear model—whether or not hidden confounders are present—where an exogenous variable is observed, without requiring an exclusion restriction.
Assuming that $(\mathbf{I} - \mathbf{B})$ is invertible, which is satisfied if the linear SCM is acyclic, we can easily express 
\begin{equation}\label{Eq:model}
    \begin{pmatrix} X \\ Y \end{pmatrix} =\mathbf{D}(\varepsilon + \mathbf{M} A),
\end{equation}
where $\mathbf{D}\in \mathbb{R}^{(d+p)\times(d+p+r)}$ existence is implied by the existence of $(\mathbf{I} - \mathbf{B})^{-1}$.

\subsection{Bound on the Perturbed Covariance} \label{Subsec:anchor_opt_space}

Let us now build upon Eq.~\eqref{Eq:model} to illustrate how distributional robustness can be formulated in a causal context and give some intuition of the proposed regularisation strategy. We show how anchor regularisation can be formulated in terms of covariances $\Sigma_{XY}$ and $\Sigma_{XY|A}$, instead of squared residuals as originally proposed in \cite{Rothenhäusler2018}, the former being more general. This relies on the simple idea that the squared residuals loss is a linear form on $\Sigma_{XY}$ and anchor regularisation simply adds a constraint on the perturbed covariance matrix $\Sigma_A$ (see Eq.~\eqref{Eq:Cgamma}).

Note that, from Eq.~\eqref{Eq:model}, we can easily express  
$\Sigma_{XY}$ as
\begin{align}\label{Eq:Sigma_split}
    \Sigma_{XY} &= \mathbf{D}\Sigma_{\varepsilon}\mathbf{D}^T + \mathbf{D}\mathbf{M}\Sigma_{A}\mathbf{M}^T\mathbf{D}^T,
\end{align}
which combines the variance from exogenous observed (anchors) and unobserved (noise) variables.

Consider the triplet $(X, Y, H) \sim \Pnu$, where $\Pnu = \mathbb{P}^{do(A \sim\nu)}_{\mathcal{C}}$ represents the perturbed distribution resulting from intervention \citep[see][Sec.~3.2]{Peters2017} on the anchor variable $A$, where the distribution of $\nu$ is assumed to be mean-centered. 
From Eq.~\eqref{Eq:Sigma_split}, the perturbed variance-covariance matrix $\Sigma_{XY}^{do(A \sim\nu)}$ can be expressed as 
\begin{equation*}
    \Sigma_{XY}^{do(A \sim\nu)} = \mathbf{D}\Sigma_{\varepsilon}\mathbf{D}^T + \mathbf{D}\mathbf{M}\Sigma_{\nu}\mathbf{M}^T\mathbf{D}^T,
\end{equation*}
with $\Sigma_{\nu} =\mathbb{E}[\nu \nu^\top]$. By imposing the constraint $\Sigma_{\nu} \preceq \gamma \Sigma_{A}$\footnote{Here, $U \preceq V$ means that $V-U$ is positive semi-definite.} 
, we can bound the interventional variance-covariance matrix $\Sigma_{XY}^{do(A \sim\nu)}$ using the training covariances $\Sigma_{\varepsilon}$ and $\Sigma_{A}$.
An intuition for this constraint is that during testing, the covariance matrix of the anchor variable can be expected to be scaled by a factor of at most $\gamma$ in all directions, thus constraining the strength of the intervention.
The anchor regulariser parameter $\gamma \in \mathbb{R}^+$ thus controls the amount of causal regularisation. This yields the specific distribution class:
\begin{align}\label{Eq:Cgamma}
    C^\gamma = \{\Pnu : \Sigma_{\nu} \preceq \gamma\Sigma_{A} \}. 
\end{align}
As demonstrated in Eq.~\eqref{Eq:Decomp_left_term} and Eq.~\eqref{Eq:Decomp_right_term} in Appendix, both $\Sigma_\varepsilon$ and $\Sigma_A$ can be expressed in terms of $\Sigma_{XY|A}$ and $\Sigma_{XY}$ of the training distribution, which leads to the following inequality:
\begin{equation}\label{Eq:train_data_bound}
    \Sigma_{XY}^{do(A \sim\nu)} \preceq \Sigma_{XY} + (\gamma-1) \Sigma_{XY|A}.
\end{equation}
Consequently, the interventional covariance matrix $\Sigma_{XY}^{do(A \sim\nu)}$ can be bounded using only the training (or unperturbed) distribution according to Eq.~\eqref{Eq:train_data_bound}.
As we will develop in \S\ref{sec:anchorcomploss}, algorithms exploiting $\Sigma_{XY}$ can thus be anchor-regularised, leading to distributionally robust estimators.
An example is the $\ell_2$ loss, defined for a one-dimensional response variable as $\mathbb{E}[\mathcal{L}(X, Y;\mathbf{b})] = \mathbb{E}[(Y - \mathbf{b}^TX)^2] = \sigma_{Y} - 2\mathbf{b}^T\sigma_{XY}+ \mathbf{b}^T\Sigma_{X}\mathbf{b}$, which is evidently linear on $\Sigma_{X Y}$. 
Similarly, the PLS algorithm \citep{Abdi2010} aims to find a pair of matrices $\mathbf{W}_x \in \mathbb{R}^{d\times u}$ and $\mathbf{W}_y \in \mathbb{R}^{p\times v}$ that maximise the covariance between the transformed data $X\mathbf{W}_x$ and $Y\mathbf{W}_y$, i.e. assuming centered $X$ and $Y$, we aim to maximise $\text{tr}(\mathbf{W}_x^T \Sigma_{XY} \mathbf{W}_y)$ while constraining the columns of $\mathbf{W}_x$ and $\mathbf{W}_y$ to each be orthogonal.
In this case, as the loss is also linear on 
matrix $\Sigma_{XY}$, i.e. PLS can be anchor-regularised, cf. \S\ref{sec:anchorcomploss}. 
These observations lead us to define a class of loss functions and demonstrate their robustness over the perturbed distributions $C^\gamma$ defined in Eq.~\eqref{Eq:Cgamma}.

\section{ROBUSTNESS THROUGH ANCHOR REGULARISATION} 
\label{sec:robustAR}

Building on the aforementioned idea, we demonstrate that anchor regularisation can be efficiently applied to a broad class of linear algorithms, including several classical MVA algorithms. This approach ensures distributional robustness for algorithms within a class $\mathbb{Q}$, which encompasses distributions arising from bounded interventions on $A$.

\subsection{\emph{Anchor-Compatible} Loss}\label{sec:anchorcomploss}
The distributional robustness properties stemming from intervention on the anchor variables arise from the ability to bound 
$\Sigma^{do(A \sim\nu)}_{XY}$ using the \emph{observational} covariance matrices $\Sigma_{XY}$ and $\Sigma_{XY|A}$. Consequently, any loss function expressed as a linear form on $\Sigma_{XY}$ can attain distributional robustness similar to AR using anchor regularisation.

\begin{definition}[\emph{Anchor-compatible} loss] \label{def:acloss}
We say that a loss function $\mathcal{L}(X, Y; \mathbf{\Theta})$ is \emph{anchor-compatible} if it can be written as $\mathcal{L}(X, Y; \mathbf{\Theta}) = f_{\mathbf{\Theta}}(C_{XY})$,
where $f_{\mathbf{\Theta}} : \mathbb{R}^{d\times p} \to \mathbb{R}$ is a linear form\footnote{A linear form is a linear map from a vector space to its field of scalars. The trace function is, for example, a linear form.} and $\mathbf{\Theta}$ are the parameters to be learned. 
\end{definition}
The following result extends Th. 1 in \citet{Rothenhäusler2018} to the previously defined class of loss functions.
\begin{theorem}
\label{thm:ACloss}
    Let the distribution of $(X, Y, H)$ be entailed by the SCM \eqref{Eq:anchor-model} and $\mathcal{L}(X, Y; \mathbf{\Theta})$ be an \emph{anchor-compatible}  loss function. Then for any set of parameters $\mathbf{\Theta}$, and any causal regulariser  $\gamma \in \mathbb{R}^+$, we have
    \begin{align}\label{Eq:theorem1}
        \begin{split}
            \sup_{\nu \in C^\gamma} \Enu[\mathcal{L}(X, Y; \mathbf{\Theta})] &= f_{\mathbf{\Theta}}(\Sigma_{X Y})
            \\
            &+ (\gamma-1) f_{\mathbf{\Theta}}(\Sigma_{X Y|A}),
        \end{split}
    \end{align}
    where 
    $C^\gamma = \{\Pnu: \Sigma_\nu \preceq \gamma \Sigma_A \}$. Proof can be found in Appendix (\ref{thm:ACloss}).
\end{theorem}
The term $f_{\mathbf{\Theta}}(\Sigma_{XY|A})$ could be understood as a causal regulariser enforcing invariance of the loss w.r.t the anchor variable. More intuitively, the anchor regularisation can be seen as adding a penalty term to how much $X$ and $Y$ covary when projected in the span of the anchor variable. The theorem suggests that in the case of distributional shifts observed during testing due to constrained intervention on $A$, the anchor-regularised estimator $\mathbf{\Theta}^{\gamma} = \argmin_{\mathbf{\Theta}} f_{\mathbf{\Theta}}(\Sigma_{X Y}) + (\gamma-1) f_{\mathbf{\Theta}}(\Sigma_{X Y|A})$ acts as an optimal estimator, as per the definition provided in Eq.~\eqref{Eq:Distrib_Robust_def}. We will proceed to demonstrate the practical application of this general result to standard MVA algorithms.
\subsection{Common Multivariate Analysis Algorithms}
%
%
\begin{table*}
    \centering
\scriptsize
\caption{Characterisation of common MVAs with anchor regularisation (loss, constraints, and anchor-compatibility).}
\label{fig:characterisationMVA}
    \begin{tabular}{lccccc}
        \toprule
         &
       {MLR}  & {OPLS}  & {RRR}  & {PLS} & {CCA}  \\
        \midrule
        Loss &
        $\| Y - \mathbf{W}^T X\|_F^2$ & $\| Y - \mathbf{U}\mathbf{V}^T X\|_F^2$  & $\| Y - \mathbf{W} X\|_F^2$ & $ -\text{tr} \left( \mathbf{W_x}^T X^T Y \mathbf{W_y} \right)$  & $-\text{tr} \left( \mathbf{W_x}^T X^T Y \mathbf{W_y} \right)$ \\
        Const. &
        - \quad & $\mathbf{U}^T\mathbf{U} = \mathbf{I}$ \quad & rank($\mathbf{W})=\rho$\quad & $\mathbf{W_x}^T\mathbf{W_x} =\mathbf{I},$ \quad & $\mathbf{W_x}^TC_X \mathbf{W_x} = \mathbf{I},$ \\
          &
         &  &   & $\mathbf{W_y}^T \mathbf{W_y}=\mathbf{I}$  & $\mathbf{W_y}^TC_Y \mathbf{W_y}=\mathbf{I}$  \\
        Comp.  &
        \cmark & \cmark &  \cmark &  \cmark &  \xmark \\
        \bottomrule
    \end{tabular}
\end{table*}

By virtue of Th.~\ref{thm:ACloss}, any \emph{anchor-compatible} loss function has theoretically grounded robustness properties to distribution shifts. This is practically relevant, as users can easily verify if a loss function is compatible using Def.~\ref{def:acloss}. In that section, we show how this can be applied to a set of commonly used multivariate algorithms in both compatible and incompatible cases. As demonstrated in the experimental section, while non-compatible loss functions might also benefit from anchor regularisation, they exhibit suboptimal robustness properties, and this regularisation should thus be used with care.

A common and straightforward approach to extend Least Squares (LS) regression to multivariate settings is treating responses independently in multioutput regression~\citep{Borchani2015}. 
This approach aims to identify the regression coefficients $\mathbf{W}$ such that $\hat{Y} = \mathbf{W}^T X$, e.g. 
by solving the LS problem $\widehat{\mathbf{W}} = \argmin_{\mathbf{W}} \| Y - \mathbf{W}^TX\|^2_F$.
When the columns of 
$Y$ exhibit correlation, RRR addresses a similar optimisation problem 
assuming that 
the rank of 
$\mathbf{W}$ is lower than $\min(d, p)$, see 
\cite{Izenman1975}. 
Alternatively, in OPLS~\citep{Arenas-Garcia1015}, we do not assume that $\mathbf{W}$ is low rank. Instead, $\mathbf{W}$ is determined by imposing the constraints $\mathbf{W} = \mathbf{V}^T\mathbf{U}$, where both $\mathbf{U}$ and $\mathbf{V}$ have a rank $\rho \leq \min(d, p)$.  
Additionally, OPLS is commonly solved through matrix deflation, eigenvalue, or generalised eigenvalue decomposition~\citep{Arenas-Garcia1015}. An advantageous property of OPLS is that the columns of the solution $\mathbf{V}$ are ordered based on their predictive performance on $Y$. This is valuable when predicting $Y$ is not the sole objective, and there is interest in learning a relevant information-retaining subspace of $X$ (here $\mathbf{V}^TX$). These are natural extensions of LS regression and are \emph{anchor-compatible}. See Appendix~\ref{sec:robustness_common_MVA} for further details and proofs. PLS and CCA are typically preferred when prioritizing learning a pertinent subspace of $X$ over predictive performance. These algoritms aim to maximise the similarity between the latent representations of predictor and target variables.  Specifically, PLS assumes the following latent representation $X = \mathbf{P}^T T + N_x$ and $Y = \mathbf{Q}^T U + N_y$ and seeks to maximise the covariance between $\mathbf{W}_x^TX$ and $\mathbf{W}_y^T Y$, while CCA aims to maximise the correlation between the estimated latent spaces. 
Correlation as a similarity measure ensures equal importance for each dimension in the learned latent space, independent of data variance. PLS, being \textit{anchor-compatible} (see Prop.~\ref{prop:ac_pls}), benefits from causal regularisation for distributional robustness. In contrast, CCA lacks anchor compatibility due to nonlinearity in its loss function concerning variance-covariance (see \S\ref{ex:incompatibility_CCA}). As illustrated in Fig.~\ref{fig:perturbation_strength}, CCA shows reduced robustness under interventions on the anchor variable's variance. Recognizing the established equivalence between CCA and OPLS \citep{Sun2009}, it may appear puzzling that one algorithm is \emph{anchor-compatible}  while the other is not. We delve deeper into this matter in the Appendix (see \S\ref{par:CCA_OPLS_eq}).

\section{TOWARDS INVARIANCE AND CAUSALITY}\label{sec:towards_causality_and_invariance}
%
%
A direct implication of Th.~\ref{thm:ACloss} is the connection between anchor regularisation and two well-studied estimators: Instrumental Variable (IV) and Partialling-Out (PA), which we 
define as follow
\begin{align}
    \mathbf{\Theta}^{IV} &= \argmin_{\mathbf{\Theta}} f_{\mathbf{\Theta}}(\Sigma_{X Y|A})  = \lim_{\gamma \to +\infty} \mathbf{\Theta}^{\gamma}\\
    \mathbf{\Theta}^{PA} &= \argmin_{\mathbf{\Theta}} f_{\mathbf{\Theta}}(\Sigma_{X Y|A}^\perp)  = \mathbf{\Theta}^{0}.
\end{align}
Here $\Sigma_{X Y|A}^\perp = \Sigma_{X Y} - \Sigma_{X Y|A}$. Note that for the MLR loss function $\mathbf{\Theta}^{IV}$ is equivalent to the two-stage least square estimation. 
A direct implication of Th.~\ref{thm:ACloss} is that the minimisers of the \emph{anchor-regularised} loss function have the following properties: unregularised parameters $\mathbf{\Theta}^1$ are optimal  for all $\nu \in C^1$, PA parameters $\mathbf{\Theta}^{PA}$ are optimal for all $\nu \in C^0$ and IV parameters $\mathbf{\Theta}^{IV}$ are optimal for all $\nu \in C^\infty$.
However, although PA and IV estimations are generally used for causal inference, this shows that they provide robustness properties when considering specific interventions on the anchor variables, even if they do not necessarily retrieve the causal parameters.
While it is known that for specific sets of interventions, causal parameters have robustness properties~\citep{Haavelmo1943TheSI}, it has recently been recognized~\citep {Christiansen2022} that this is not always the case. The following proposition shows the connection between the optimal structural parameters in terms of error and the causal parameters.
\begin{proposition}
    Let us denote \( R(\tilde{\mathbf{B}}) = [X, Y, H]^T - \tilde{\mathbf{B}}[X, Y, H]^T \) as the error term when reconstructing \((X, Y, H)\) with \(\tilde{\mathbf{B}}\). Given any matrix \(\Bmax\) such that
 
    \begin{align*}
        \begin{split}
            \sup_{\nu \in C^{\gamma}} \Enu [ f_{\mathbf{\Theta}}( R(\Bmax) &\otimes R(\Bmax) )] \\
        &\leq \sup_{\nu \in C^{\gamma}} \Enu [ f_{\mathbf{\Theta}}( R(\mathbf{B}) \otimes R(\mathbf{B}) )],
        \end{split}
    \end{align*}
    it holds that if the loss $f_{\mathbf{\Theta}}$ is \emph{anchor-compatible}, then
    \begin{align}
        \begin{split}\label{eq:suboptimality_causal_param}
            \sup_{\nu \in C^{\gamma}} f_{\mathbf{\Theta}}( &(\mathbf{B} - \Bmax) \Sigma_{XYH}^{do(A \sim\nu)} (\mathbf{B} - \Bmax)^T) \\
       &\leq 4 f_{\mathbf{\Theta}}(\Sigma_\epsilon) + 4 \gamma f_{\mathbf{\Theta}}(\mathbf{M}\Sigma_A \mathbf{M}^T).
        \end{split}
    \end{align}
\end{proposition}

The worst case risk between the causal parameter $\mathbf{B}$ and the optimal one $\Bmax$ is bounded but increase linearly with $\gamma$. Thus the potential improvement enabled by using \emph{anchor-regularisation} instead of seeking to retrieve causal parameters becomes clearer as intervention strength on the anchor increases.  

\section{ESTIMATORS}\label{sec:estimators}

In the previous sections, we derived properties of a regularised version of various MVA methods in the population case. We consider now the sample setting where we are given $n$ i.i.d observations of $(X, Y, A)$.  
Observations are collectively organised row-wise, in the following matrices,
$\mathbf{X} \in\mathbb{R}^{n \times d}$, $\mathbf{Y} \in\mathbb{R}^{n \times p}$ and $\mathbf{A} \in\mathbb{R}^{n \times q}$. 
For a given \emph{anchor-compatible} loss $\mathcal{L}(X,Y;\mathbf{\Theta}) = f_{\mathbf{\Theta}}(C_{XY})$, we propose a simple estimator of the population parameters:
\begin{equation}\label{Eq:estimator_full}
    \widehat{\mathbf{\Theta}}^\gamma = \argmin_{\mathbf{\Theta}} f_{\mathbf{\Theta}}(\mathbf{S}_{XY}) + (\gamma-1) f_{\mathbf{\Theta}}(\mathbf{S}_{XY|A}),
\end{equation}
where $\mathbf{S}_{XY}$ and $\mathbf{S}_{XY|A}$ are respectively the empirical estimators of $\Sigma_{XY}$ and of $\Sigma_{XY|A}$. 


\paragraph{Parameter Selection}

When knowledge about the graph and anchor variable distributional shift is available, one can estimate the optimal $\gamma$ for robustness against worst-case scenarios. For instance, if the expected anchor variable perturbation strength is up to $1.5$, \citet{Rothenhäusler2018} recommends $\gamma=1.5$. 
Without prior knowledge, various strategies can be considered: setting $\gamma$ to a low value (e.g. $\gamma=2$) might be a suitable default choice, while \citet{Sippel2021} and \citet{Szekely2022} suggest selecting $\gamma$ as a trade-off between prediction error (MSE or $R^2$ score) and the correlation between residuals and the anchor variable, or the value of projected residuals in the anchor variable's span. Similar strategies have been proposed in deemed related algorithms~\citep{cortes2022physics,li2022kernel}.


\begin{figure*}[t!]
    \centering
    \includegraphics[width=\textwidth]{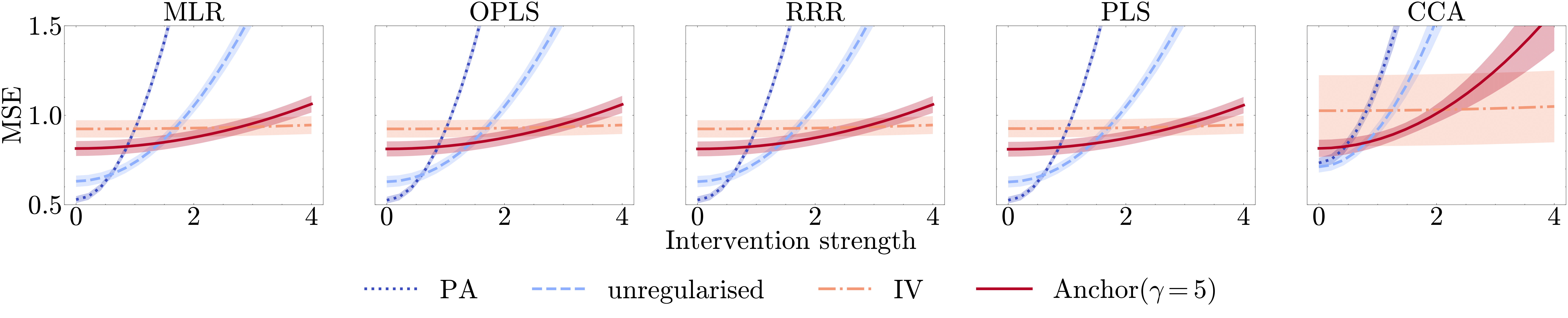}
    \vspace{-0.45cm}
    \caption{Robustness to increasing perturbation strength for PA, unregularised, IV, anchor-regularised ($\gamma=5$)  algorithms ( MLR, OPLS, RRR, PLS and CCA). Anchor versions show robustness in terms of $R^2$ for bounded intervention strength. Shaded areas represent the range of two standard errors of the mean for running $B=20$ times the experiment.}
    \label{fig:perturbation_strength}
\end{figure*}

\paragraph{High Dimensional Estimators}

Dealing with high dimensional data challenges estimators and often requires introducing regularisation~\citep{buhlmann2011statistics, candes2007dantzig}. Regularisation has been generally introduced to control models' capacity and avoid overfitting~\citep{girosi1995regularisation,Tibshirani1996,hastie2009elements}, but also to introduce prior (domain) knowledge in the algorithms, as in AR. 
Adding a regularisation term $\Omega(\|\mathbf{\Theta}\|)$ to the empirical loss in Eq.~\eqref{Eq:estimator_full} leads to estimate $\mathbf{\Theta}$ as
$\argmin_{\mathbf{\Theta}} f_{\mathbf{\Theta}}(\mathbf{S}_{XY})+ (\gamma-1) f_{\mathbf{\Theta}}(\mathbf{S}_{XY|A}) + \Omega(\|\mathbf{\Theta}\|)$.

In CCA, OPLS, and PLS regression (or the rank in RRR), the number of components can be viewed as regularisation hyperparameters. 
Consequently, the optimisation task for anchor-regularised MVA algorithms may involve many hyperparameters. For instance, in Anchor-regularised Reduced Rank Ridge Regression (RRRR)~\citep{Mukherjee2011}, three hyperparameters require tuning: $\ell_2$ regularisation $\alpha$, rank $\rho$, and anchor regularisation $\gamma$. Since each hyperparameter optimization addresses different objectives, it may be impractical to optimise all of them simultaneously. 

\paragraph{Limitations}\label{par:limit}
Since anchor regularisation generally incurs only a minor additive computational cost (see Appendix \S\ref{sec:compexity}), we believe it could be used in a variety of fields to leverage robustness guarantees. However, a few limitations should be kept in mind when considering its use.
First, it should be clear to the reader that \emph{anchor-regularisation} is primarily designed for distributional robustness, not for recovering causal parameters. Also, the data distribution must be entailed in SCM Eq.~\eqref{Eq:model},  which also entails linearity assumptions; without this, robustness guarantees are invalid. Lastly, the choice of the regularisation parameter $\gamma$ is crucial for optimal performance, but finding a sensible value can be challenging in some contexts.

\section{EXPERIMENTS}
Our theoretical findings are substantiated through an extensive series of experiments that highlight the robustness properties of \emph{anchor-regularised} MVA algorithms. 
Furthermore, we provide a high-dimensional example to elucidate the process of hyperparameter selection, and demonstrate its practical relevance by showcasing how anchor regularisation enhances climate predictions in a real-world climate science application.
\subsection{Simulation Experiments}\label{Par:Toy_model_results}
To demonstrate how anchor regularisation can be applied with different multivariate analysis algorithms, we adapt the experiments from \citet{Rothenhäusler2018} to a multioutput setting.
In particular, we assume that the training data $(A, X, Y)$ follows a distribution entailed by the following linear SCM
\begin{align}\label{Eq:toy_model_IV}
    \begin{split}
        \varepsilon_A, \varepsilon_H, &\varepsilon_X, \varepsilon_Y \sim \mathcal{N}(0, 1)\\
        A &\leftarrow \varepsilon_A\\
        H &\leftarrow \varepsilon_H\\
        X &\leftarrow A\mathbf{1}_p^\top  + H \mathbf{1}_p^\top + \varepsilon_X\\
        Y &\leftarrow \mathbf{W}^TX + H \mathbf{1}_d^\top + \varepsilon_Y,
    \end{split}
\end{align}
 where $X$ and $Y$ are of dimension $d=p=10$ ($d=p=300$ with $n=200$ in the high dimensional setting), $\mathbf{1} \in \mathbb{R}^d$ is a ones vector, and $\mathbf{W}$ is low rank $\rho$ (see Appendix \S\ref{par:toy_experimental_setting} for more details on the experiments settings). We generate test data by intervening on the anchor variable's distribution, setting $A\sim \mathcal{N}(0, t)$, where $t$ is the \emph{perturbation strength}. Each MVA algorithm assumes oracle knowledge of the rank $\rho$ of $\mathbf{C}$, aligning RRR's rank, PLS regression's component count, and CCA accordingly. We consider an
 IV setting ($A$ only affects $Y$ through $X$) and we show that even in this scenario, anchor regularised algorithms (with $\gamma=5$) outperform the PA, IV, and unregularised algorithms.
Shown in Fig.~\ref{fig:perturbation_strength}, all \emph{anchor-compatible} algorithms exhibit robustness to distribution shifts. Anchor-regularised models (with $\gamma=5$) excel with bounded-strength interventions; PA regularisation is optimal for weak interventions; and IV regularisation for unlimited perturbation strength. Overall, anchor-regularised algorithms maintain stable performance across various perturbation strengths. 
We also conducted a set of high-dimensional experiments ($p\gg n$ and $d\gg n$) using RRR and MLR both regularised with $\ell_2$ norm (ridge regularisation). In both cases, the anchor-regularised algorithms exhibit robustness to increasing perturbation strength (see Fig.~\ref{fig:perturbation_strength_high_dim} in \S\ref{par:toy_experimental_setting}). This is also the case when $A$ is a confounder (affecting both $X$ and $Y$, potentially through $H$), in which case anchor-regularised algorithms present are equally optimal for a wide range of perturbation strength (see Fig.~\ref{fig:fig3_anchor_paper} in \S\ref{sec:confounding_experiment}). Though not as pronounced as in \emph{anchor-compatible} algorithms, anchor-regularised CCA also displays robustness to perturbations in $A$, prompting further investigation into its behaviour.

\label{sec:high_dim_sim_exp}
We showcase how hyperparameter selection can be performed through a simulation experiment in a high-dimensional setting using an anchor-regularised RRRR, an $\ell_2$-regularized version of RRR with $\alpha$ the ridge hyperparameter and $\rho$ the reduced rank.
In this setting, we are given three hyperparameters:  $\gamma$ enforces robustness to intervention on the anchor, and $\alpha$ and  $\rho$ aim to maximise prediction performance. Following~\citep{Sippel2021} we select hyperparameters as a trade-off between predictive performance (measured via MSE) and correlation between anchor and residuals. We give equal weights to both objectives but this can be adapted regarding the knowledge available and the application. Thus hyperparameters are selected at testing such that they minimise the combination of the two objectives.
Fig.~\ref{fig:pareto-ARRRR} shows how augmenting regularisation for distributional robustness to anchor intervention often reduces predictive performance in the training sample (Fig.~\ref{fig:pareto-ARRRR}.A), yet the trade-off translates into improved predictive performance in testing samples.
More details can be found in~\S\ref{par:toy_high_dim_detail}.

\subsection{Robust Climate Prediction}\label{sec:realclimate}
We showcase the efficacy of our approach in a real-world application within the Detection and Attribution of Climate Change (D\&A) domain. We extend the methodology of~\citet{Sippel2021}, who utilised AR for robust detection of forced warming against increased climate variability,
by applying it to predict multidimensional local climate responses. Given the increased variability observed in recent climate models~\citep{Parsons2020} and observations~\citep{Desole2006, Kociuba2015, Anson2017}, our approach aims to ensure robustness against potential underestimation of decadal and multidecadal internal climate variability \citep{Parsons2020, Mcgregor2018}.
\paragraph{Objective}
We use a $p$-dimensional temperature field $X_{\text{mod}}$ to predict a temperature response $Y_{\text{mod}}^{\text{forced}}\in \mathbb{R}^p$ to external forcings (such as greenhouse gas emissions, aerosols, solar radiation, or volcanic activity), a crucial step in detecting warming using the fingerprint method \citep[see][]{Hegerl1996}, while ensuring robustness to climate's Decadal Internal Variability (DIV). This leads to the following linear model $Y_{\text{mod}}^{\text{forced}} = X_{\text{mod}} \beta + \epsilon.$
We follow
\citet{Sippel2021}, leveraging multiple climate models from the Climate Model Intercomparison Project (CMIP), both Phase 5 \citep{Taylor2011} and 6 \citep{Eyring2016} (CMIP5 and CMIP6). Specifically, we employ four models (CCSM4, NorCPM1, CESM2, HadCM3) characterised by lower-scale DIV to train our MVA algorithm and validate its robustness using anchor regularisation against models exhibiting higher-magnitude DIV (CNRM-CM6-1, CNRM-ESM2-1, IPSL-CM6A-LR). 

\begin{figure*}[t!]
    \centering
    \begin{minipage}[t]{0.3\textwidth}
        \centering
        \includegraphics[width=\textwidth]{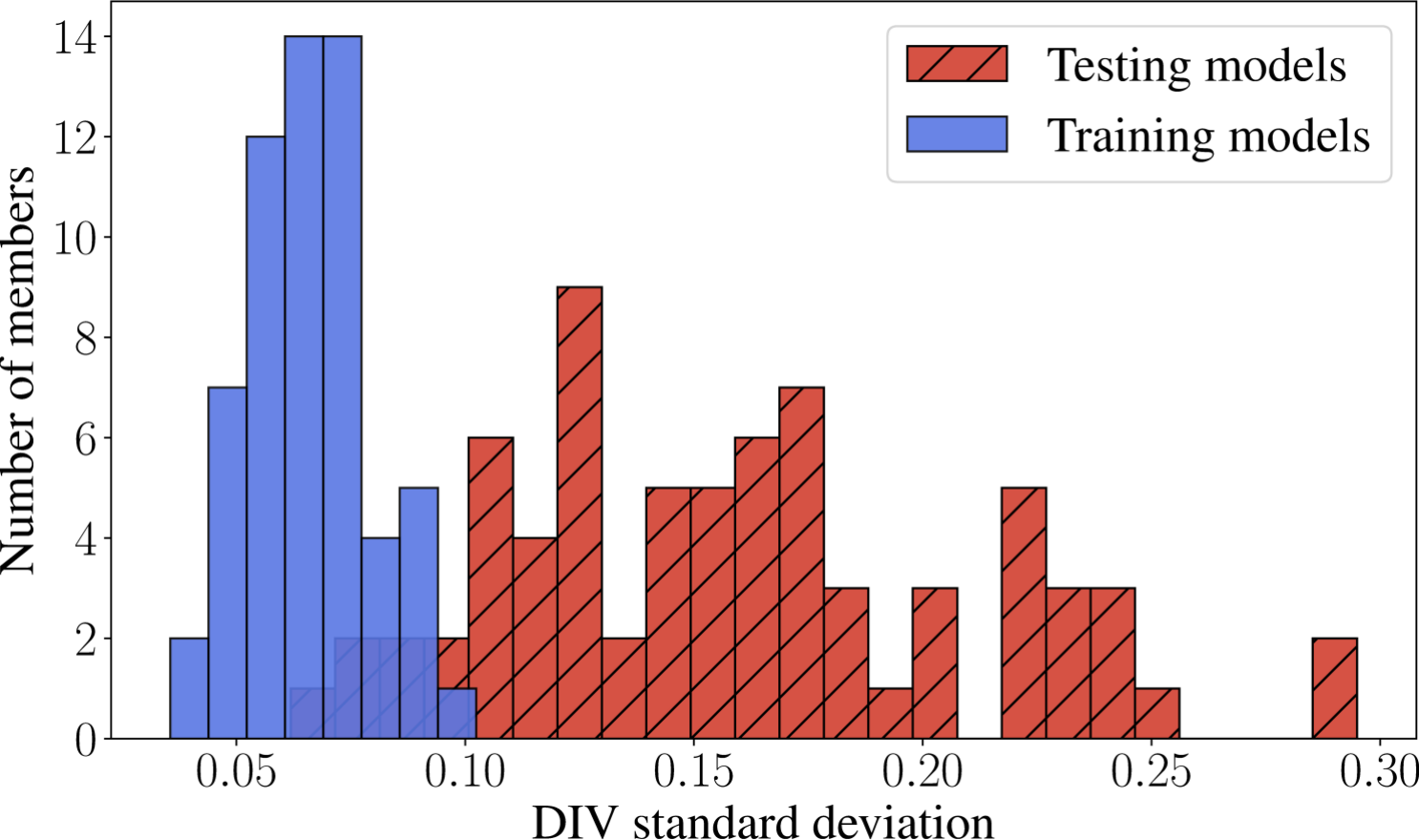}
        \vspace{-0.45cm}
        
    \end{minipage}
    \hfill
    \begin{minipage}[t]{0.65\textwidth}
        \centering
        \includegraphics[width=\textwidth]{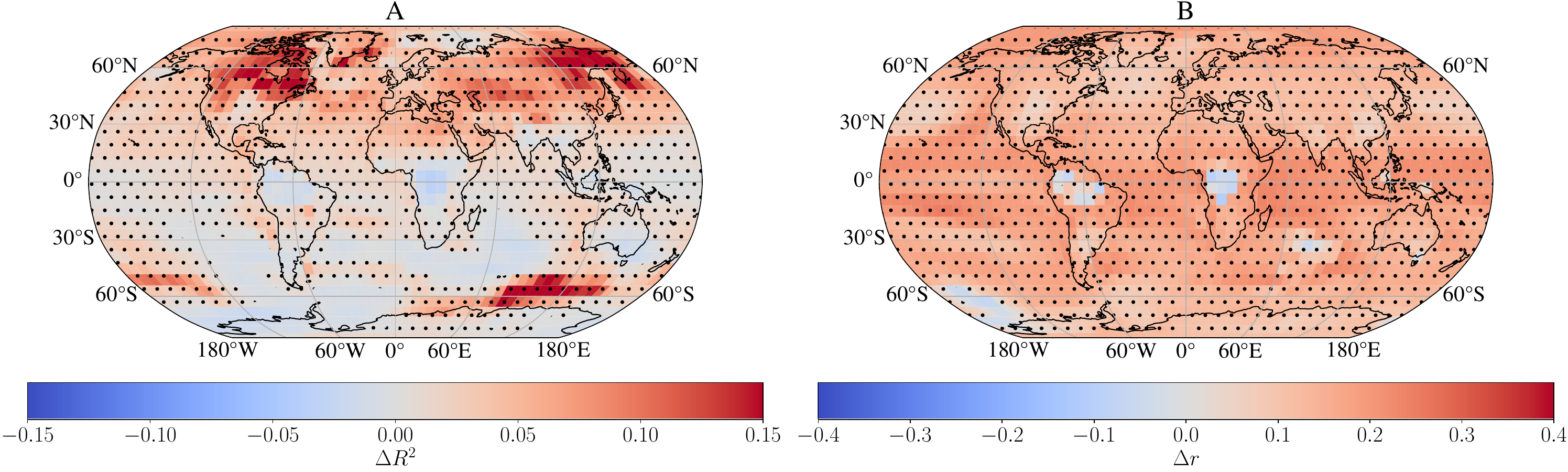}
        \vspace{-0.25cm}
        
        \vspace{-0.5cm}
    \end{minipage}
    \caption{(Left) Standard deviation of DIV among training (red) and testing (blue) model members. (Right A) $R^2$ score differences between A-RRRR ($\gamma=5$) and RRRR for test models. (Right B) Differences in residuals-DIV correlation ($r$) between A-RRRR ($\gamma=5$) and RRRR for test models. Red and hatched areas indicate where A-RRRR performs better.}
    \label{fig:combined}
\end{figure*}

\paragraph{Estimators}
We employ RRRR as MVA algorithm. This choice is motivated by the correlation structures present in both the predictors (temperature fields) and the target (temperature response to external forcing), which arise due to spatial autocorrelation. We use DIV as anchor to protect against shifts in long term internal variability. 
As both $\alpha$ and $\rho$ in RRRR serve to regularise the regression, we optimise them using cross-validation across the training models. We consider two levels of anchor regularisation: $\gamma=5$ (low regularisation) and $\gamma = 100$ (high regularisation) to showcase how various regularisation strategies lead to different estimated forced response. We evaluate our results regarding two metrics: $R^2$ score (a standard metric when predicting spatial fields) and mean correlation between the anchor (DIV) and the regression residuals noted as $r$, similarly to what is done in~\citep{Sippel2021, Szekely2022}.
\paragraph{Data}
We selected $7$ models from the CMIP5 and CMIP6 archives, each containing at least $8$ members from historical simulations, to ensure accurate estimation of the climate response (refer to Tab.~\ref{tab:CMIP56_models} for detailed model information). The data preprocessing procedure for each model involves re-gridding surface air temperature data to a regular 
\ang{5}$\times$\ang{5} grid and computing yearly anomalies by subtracting the mean surface air temperature for the reference period: years $1850-1900$. The forced response $Y^{\text{forced}}_{\text{mod}}$ is obtained using a standard approach \citep[see][and Appendix \ref{sec:real_world_details}]{Deser2020}, averaging over all available members in each model. Furthermore, we use DIV as a proxy for multidecadal climate internal variability \citep[see][]{Parsons2020}, achieved by removing the global forced response to global temperature and smoothing it using a $10$-year running mean. This procedure is commonly employed for estimating the multidecadal variability of the climate system \citep[see][for further details]{Sippel2021, Deser2020}. Finally, all data are standardised.

\paragraph{Evaluation Procedure}

We categorise the models into two groups (low and high variability) based on the distribution of the standard deviation of their DIV among the members of each model. 
As depicted in Fig.~\ref{fig:combined} (Left), a significant proportion of members used for training display a notably smaller magnitude of DIV.
Hyperparameters $(\alpha, \rho)$ are selected through a leave-half-models-out cross-validation procedure. Given $k=4$ training models, we randomly select $2$ models for training and $2$ models for validation, and repeated $B=10$ times. We use $500$ samples from each model to ensure equal weights are given to each training model in the learning algorithm, despite variations in the number of members. For each of the $B$ train/validation splits, we train an RRRR model and select the hyperparameters that yield the highest averaged $R^2$ score across the $B$ sampling splits. $R^2$ provides a performance measure comparable across regions and is typically used when predicting spatial variables \citep[see e.g.][]{Sippel2019}. Hyperparameter $\alpha$ is selected from $20$ candidates in a logarithmically-spaced sequence ($\lambda \in [1; 10^6]$), and $\rho$ is selected from $21$ candidates in a linearly-spaced integer sequence ($\rho \in [300; 600]$).
\paragraph{Results}
For both anchor-regularised RRRR (A-RRRR) with $\gamma=5$ and $\gamma=100$, we observe an improvement in both metrics in the testing set (respectively $0.537$ and $0.533$ of $R^2$ for A-RRRR and $0.506$ for unregularised RRRR), while experiencing a slight decrease of performance in the training samples ($0.510$ of $R^2$ for unregularised RRRR and respectively $0.500$ and $0.487$ for both A-RRRR). As the latter strongly protects against shifts in DIV, we notice that its mean correlation between residuals and DIV is lower, albeit with a slight decrease in the $R^2$ score (see Table~\ref{tab:metrics_rrrr} in Appendix \S\ref{sec:real_world_details}). 
In Fig.~\ref{fig:combined}, we observe that anchor-regularised RRRR performs better in the northern hemisphere, particularly in regions where the residuals are highly correlated with DIV and in the northern hemisphere where unregularised RRRR performs poorly (see Figs.~\ref{fig:R2_corr_plot_gamma100} and~\ref{fig:R2_corr_plot_rrrr} in Appendix \S\ref{sec:real_world_details}). Conversely, we observe a decrease in the performance of A-RRRR in regions where unregularised RRRR already performs well or where residuals are anti-correlated with DIV. The latter challenge might be addressed by considering regional proxies of internal variability instead of a global one, which does not always accurately represent local internal variability.

These results suggest the potential of using anchor-regularised multi-output algorithms in D\&A studies to detect and attribute local responses to external forcings, a fundamental problem in climate science.

\subsection{Robust Air-Quality Prediction}\label{sec:realair}

We conduct an experiment to evaluate the effectiveness of anchor regularization in ensuring robust predictions under temporal distribution shifts in pollution variables. Our approach is compared against established robust prediction methods.

\begin{table*}[t]
    \centering
    \small
    \caption{Mean, Median, Max, and Min MSE values for different models, along with the number of cases where they performed better than LR. IRM shows a better worst-case risk control but A-PLS has a lower median MSE and has better performances than LR in 22 out of the 24 cases.}
    \begin{tabular}{lccccc}
        \toprule
        Model & Mean MSE & Median MSE & Max MSE & Min MSE & Better than LR \\
        \midrule
        {IRM}    & \textbf{0.6852}  & 0.7297  & \textbf{0.8655}  & \textbf{0.4580}  & 20 \\
        {CVP}    & 0.8171  & 0.7907  & 1.1872  & 0.5764  & 21 \\
        {Ridge}  & 0.7699  & 0.7518  & 0.9969  & 0.5525  & 17 \\
        {LR}     & 0.8213  & 0.7914  & 1.2146  & 0.5785  & -  \\
        {AR}     & 0.7804  & 0.7406  & 1.2146  & 0.5058  & 16 \\
        {A-Ridge} & 0.7809  & 0.7627  & 0.9969  & 0.5877  & 13 \\
        {A-PLS}  & 0.7231  & \textbf{0.7251}  & 0.9931  & 0.4788  &\textbf{22} \\
        \bottomrule
    \end{tabular}
    \label{tab:mse_comparison}
\end{table*}
\paragraph{Objective}
We assess the performance of anchor regularization in handling temporal shifts by testing models across different seasons, where the relationship between meteorological variables and air quality indicators is known to vary.
\paragraph{Data}
We use the \href{https://archive.ics.uci.edu/dataset/360/air+quality}{Air Quality dataset} from \cite{air_quality_360}, leveraging meteorological variables (Temperature, Humidity, and Relative Humidity) as predictors and 10 air quality indicators as outcomes (see Table \ref{tab:vairable_pollution} in Appendix). Seasons are treated as categorical variables, and models are tested on unseen seasons to evaluate their robustness under temporal shifts.
\paragraph{Estimators}
Several linear model-based estimators are compared. Invariant Risk Minimization (IRM) follows the approach of \cite{Arjovsky2019}. Conditional Variance Penalties (CVP), as proposed by \cite{heinze2021conditional}, employing quantile binning for outcome discretization. We also consider Linear Regression (LR) and Ridge Regression (Ridge) as baseline models. Finally, we include Anchor Multioutput Regression (AR) and Anchor PLS Regression (A-PLS) to assess the impact of anchor regularization.
\paragraph{Evaluation Procedure}
Models are trained on two seasons, validated on a third for hyperparameter tuning, and tested on the fourth, cycling through all 24 possible combinations. The hyperparameters considered include anchor regularization, selected from a logarithmically-spaced sequence ($\gamma \in [10^{-10}; 10^{-10}]$); the number of PLS components, varying between one and three; IRM and CVP regularization parameters, selected from a logarithmic grid ($\lambda \in [10^{-3}; 1]$); and Ridge regularization, also selected from a logarithmic grid ($\lambda \in [10^{-10}; 10^{-10}]$). All hyperparameters are selected from 20 candidates.
Mean Squared Error (MSE) is used as the primary evaluation metric. We report the mean, median, maximum, and minimum MSE across the 24 combinations, as well as the number of splits where a model outperforms LR in mean MSE.

\paragraph{Results}
All out-of-distribution methods—IRM, CVP, AR, and A-PLS—demonstrate superior performance compared to linear regression (LR). A-PLS shows competitive results alongside IRM, surpassing LR in 22 out of 24 combinations and achieving better median MSE. However, its slightly higher mean and maximum MSE suggest that IRM remains the preferred choice for minimising worst-case risk. Still, anchor-regularised PLS exhibits significantly better control over worst-case risk (Max MSE) compared to simple multi-output anchor regression, highlighting the potential of anchor regularisation to leverage advanced multivariate analysis techniques in practical, real-world scenarios.


\section{CONCLUSION}\label{sec:discussion}

In this study, we extend the causal framework of anchor regression proposed by \citet{Rothenhäusler2018}, demonstrating the versatility of their regularisation approach across a wide range of MVA algorithms. Given the significant challenge of generalising models to OOD data in machine learning, we advocate for integrating anchor regularisation into a broader class of MVA algorithms, including RRR, OPLS, or PLS regression, particularly when domain knowledge suggests a bounded intervention strength on the anchor.
Moreover, we highlight that anchor regularisation offers an interesting trade-off between prediction performance and invariance to distribution shifts, addressing concerns regarding over-conservativeness in some cases.
Future theoretical advancements will entail extending the formulation 
nonlinear cases using kernel methods. Furthermore, an interesting avenue for exploration is understanding how statistical learning algorithms incompatible with anchor regularisation behave when this regularisation is applied. 
On the application front, we explore how anchor regularisation, when combined with various MVA algorithms, can be used to detect forced responses to more complex climate variables (e.g., precipitation, temperature, and their extremes) and attribute these responses to external forcing sources, such as greenhouse gas or aerosol emissions (anthropogenic factors), or natural phenomena like solar radiation and volcanic activity. Additionally, we demonstrate in an air quality prediction problem that, in the context of categorical anchors, anchor-regularised MVA yields results that are competitive with state-of-the-art approaches, such as IRM.
We anticipate a broad development and application of anchor-regularised methods across various fields of science.

\subsubsection*{Acknowledgements}
Paper accepted in Artificial Intelligence and Statistics 2025 conference (AISTATS 2025).

Authors acknowledge funding from the Horizon project AI4PEX (grant agreement 101137682), the Horizon project ELIAS (grant agreement 101120237), and the European Research Council (ERC) support under the ERC Synergy Grant USMILE (grant agreement 855187).

\newpage

\bibliographystyle{unsrtnat} 
\bibliography{references2}

\appendix

\input{supp}

\end{document}

%% file: supp.tex
\clearpage
\newpage

\onecolumn
\appendix
\section{ROBUSTNESS OF COMMON MULTIVARIATE ANALYSIS ALGORITHM}\label{sec:robustness_common_MVA}
In this section, we aim to prove the compatibility and incompatibility of some standard multivariate analysis algorithms. 
We assume that the $(X, Y, H, A)$ distribution be entailed in Eq.~\eqref{Eq:anchor-model}.

A natural extension of Ordinary Least Squares ($\ell_2$) regression to the multioutput case is to solve an OLS problem for each output, which is equivalent to solving the optimisation problem:
\begin{align}
    \min_{\mathbf{W} \in \mathbb{R}^{d\times p}} \mathbb{E}\|Y - \mathbf{W}^TX\|_F^2,
\end{align}
also known as multilinear (or multioutput) regression. Extra constraints on the regression coefficients can be added, leading, for example, to Reduced Rank Regression by constraining $\mathbf{W}$ to be of rank $\rho < \min(p, d)$. This constraint resembles constraining $\mathbf{W}$ to have the form $\mathbf{V}^T\mathbf{U}$, with $\mathbf{U} \in \mathbb{R}^{\rho\times d}$ and $\mathbf{V}\in \mathbb{R}^{\rho\times p}$.

This formulation is closely related to the formulation of Orthogonalised Partial Least Squares with an extra constraint on the ordering of the columns of $\mathbf{V}^TX$ based on their predictive performance on $Y$. Therefore, we derive the proof of anchor compatibility for the multilinear regression setting, which directly extends to anchor compatibility of OPLS and Reduced Rank Regression.

Since both OPLS and RRR aim to achieve the same objective as defined in Eq.~\eqref{Eq:OPLS_RRR} with different constraints on regression coefficients $\mathbf{W}$, we prove the general anchor compatibility of MLR, but this extends to OPLS and RRR.

\begin{proposition}[Multilinear, Reduced Rank and Orthonormalised Partial Least Square Regression are \emph{anchor-compatible}.]
    The algorithm minimising the expectation of the following loss function:
    \begin{align}\label{Eq:OPLS_RRR}
        \mathcal{L}(X, Y; \mathbf{\Theta}) = \|Y - \mathbf{W}^TX\|^2_F
    \end{align}
    is \emph{anchor-compatible}. Here $\|.\|_F$ defines the Froebinius norm.
\end{proposition}

\begin{proof}
The proof of this proposition is straightforward, observing that the loss can be written as:
\begin{align}
    \mathcal{L}(X, Y; \mathbf{\Theta}) = \sigma_Y - \mathbf{W}^T\sigma_{XY} + \mathbf{W}^T\sigma_{X}\mathbf{W},
\end{align}
which is linear over the variance-covariance $\Sigma_{XY}$.

\end{proof}

Another type of Multivariate Analysis algorithm is not aimed at minimising prediction error. Instead, it focuses on maximising the similarity between latent representations of the predictors $X\mathbf{W}_x$ and the target $Y\mathbf{W}_y$. A standard measure of this similarity is covariance, leading directly to the Partial Least Squares (PLS) algorithm, for which we now present anchor compatibility.

\begin{proposition}[Partial Least Square regression is \emph{anchor-compatible}.]\label{prop:ac_pls}
    The PLS Regression algorithm maximising the expectation of the following loss function is \emph{anchor-compatible}:
    \begin{align}\label{Eq:PLS_loss}
        \mathcal{L}(X, Y; \mathbf{W}) = \frac{\text{tr}(\mathbf{W_x}^TC_{XY}\mathbf{W_y})}{\sqrt{\text{tr}(\mathbf{W_x}^T\mathbf{W_x})}\sqrt{\text{tr}(\mathbf{W_y}^T\mathbf{W_y})}}.
    \end{align}
     
\end{proposition}

\begin{proof}
The loss function Eq.~\ref{Eq:PLS_loss} is clearly linear with respect to the variance-covariance $\Sigma_{XY}$, which is sufficient to ensure the anchor compatibility of PLS regression.
\end{proof}

On the other hand, one could consider correlation as a measure of the similarity between the learned latent spaces, aiming to account for potential differences in the variance of each variable and thus assigning equal weight to each dimension of the latent space. This approach is generally known as Canonical Correlation Analysis (CCA) or PLS mode B. By considering the variance of the latent representation of the predictors and the target, we sacrifice linearity with respect to the variance-covariance of $X$ and $Y$ (as illustrated in Eq.~\ref{Eq:CCA_loss}), making CCA incompatible with anchor regularisation.

\begin{example}[Canonical Correlation Analysis is not \emph{anchor-compatible}.]\label{ex:incompatibility_CCA}
    The Canonical Correlation Analysis solving the optimisation problem 
    \begin{align}\label{Eq:CCA_loss}
        \mathcal{L}(X; \mathbf{W}) = \frac{\text{tr}(\mathbf{W_x}^T\Sigma_{XY}\mathbf{W_y})}{\sqrt{\text{tr}(\mathbf{W_x}^T\Sigma_{X}\mathbf{W_x})} \sqrt{\text{tr}(\mathbf{W_y}^T\Sigma_{Y}\mathbf{W_y})}}
    \end{align}
    is not \emph{anchor-compatible} as it is not linear over the variance-covariance matrix. This explains the different behaviour taken by anchor regularisation in Fig.~\ref{fig:perturbation_strength} by the anchor-regularised CCA.
\end{example}

This is illustrated in Fig.~\ref{fig:perturbation_strength}, where we can observe that anchor-regularised CCA exhibits a distinct behavior compared to anchor-compatible MVA algorithms. It would be of interest to investigate to which extent the incompatibility of CCA with anchor regularisation impacts its distributional robustness properties and if its anchor regularisation could still be of interest.

\paragraph{OPLS Formulations and their Relation to CCA}\label{par:CCA_OPLS_eq}
There are two versions of OPLS: the standard eigenvalue decomposition (EVD) and the generalized eigenvalue (GEV) formulations~\cite{Arenas-Garcia1015}. In our work, we implement EVD-OPLS whose optimization problem is 
\begin{align*}
\min_{\mathbf{U},\mathbf{V}} & \| Y - \mathbf{U} \mathbf{V}^T X\|_F^2\\
&\text{s.t. } \ \mathbf{U}^T\mathbf{U} = \mathbf{I}.
\end{align*}
EVD-OPLS is linear in $C_{XY}$ and is, therefore, an anchor-compatible loss. On the other hand, the GEV-OPLS loss is 
\begin{align*}
\min_{\mathbf{U},\mathbf{V}} & \| Y - \mathbf{U} \mathbf{V}^T X\|_F^2\\
&\text{s.t. } \ \mathbf{V}^TC_X\mathbf{V} = \mathbf{I} 
\end{align*}
Viewing the optimization for $\mathbf{V}$ as a generalized eigenvalue decomposition problem by absorbing the constraint into the loss gives us the equivalent optimization 
\begin{equation*}
    \min_{\mathbf{v}} \frac{\mathbf{v}^T X^T Y Y^T X \mathbf{v}}{\mathbf{v}^T C_X \mathbf{v}}.
\end{equation*}
This is clearly \emph{not} linear in $C_{XY}$ because it is not linear in $C_{X} = X^TX$.

Furthermore, GEV-OPLS and CCA are shown to be equivalent up to an orthogonal rotation \citep[see][Theorem 2]{Sun2009}, and by similar reasoning, CCA is also not anchor-compatible.

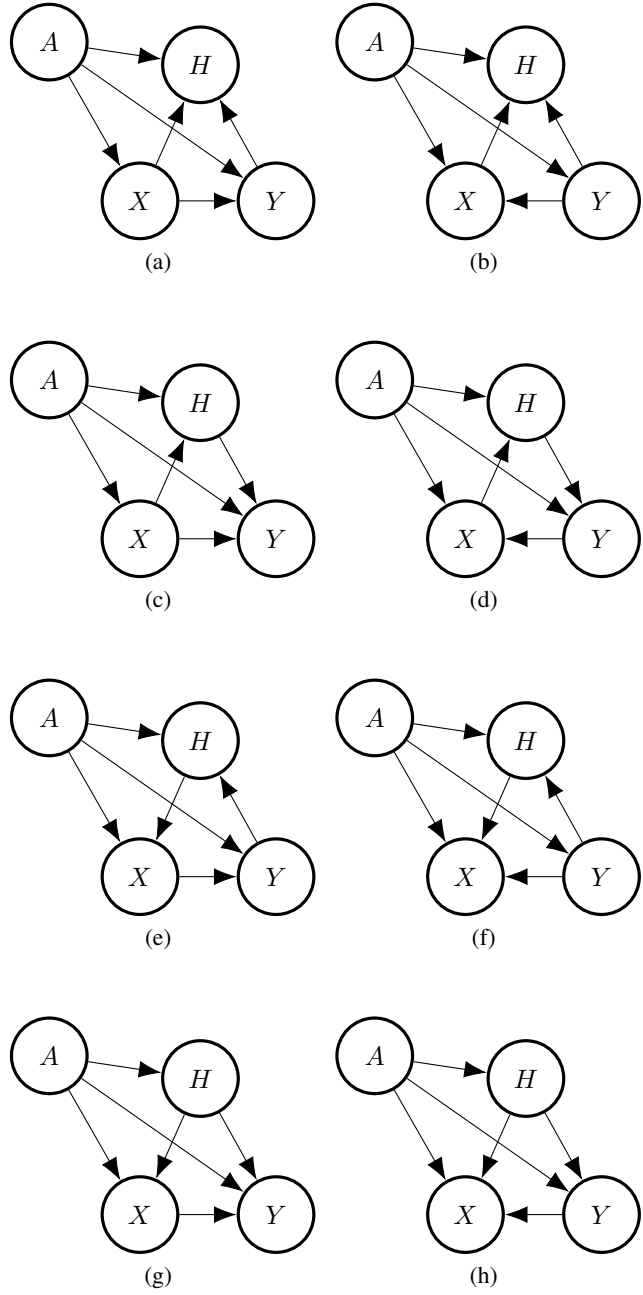
\begin{figure}
    \centering
    \subfigure[]{
        \begin{tikzpicture}[auto, scale=1,
            node distance = 18mm and 6mm,
            normaledge/.style ={arrows=-{Latex[length=3mm]}},
            node/.style={circle,inner sep=1mm,minimum size=1cm,draw,black,very thick,text=black}]
            \node [node] (x) {$X$};
            \node [node] (y) [right of = x] {$Y$};
            \node [node] (h) [above of = x, xshift=8mm]  {$H$};
            \node [node] (a) [above of = x, xshift=-12mm, yshift=3mm]  {$A$};
            \path[normaledge] (a) edge (x);
            \path[normaledge] (a) edge (y);
            \path[normaledge] (a) edge (h);
            \path[normaledge] (x) edge (y);
            \path[normaledge] (y) edge (h);
            \path[normaledge] (x) edge (h);
        \end{tikzpicture}
    }
    \subfigure[]{
        \begin{tikzpicture}[auto, scale=1,
            node distance = 18mm and 6mm,
            normaledge/.style ={arrows=-{Latex[length=3mm]}},
            node/.style={circle,inner sep=1mm,minimum size=1cm,draw,black,very thick,text=black}]
            \node [node] (x) {$X$};
            \node [node] (y) [right of = x] {$Y$};
            \node [node] (h) [above of = x, xshift=8mm]  {$H$};
            \node [node] (a) [above of = x, xshift=-12mm, yshift=3mm]  {$A$};
            \path[normaledge] (a) edge (x);
            \path[normaledge] (a) edge (y);
            \path[normaledge] (a) edge (h);
            \path[normaledge] (x) edge (h);
            \path[normaledge] (y) edge (h);
            \path[normaledge] (y) edge (x);
        \end{tikzpicture}
    }
    
    \vskip 0.5cm
    
    \subfigure[]{
        \begin{tikzpicture}[auto, scale=1,
            node distance = 18mm and 6mm,
            normaledge/.style ={arrows=-{Latex[length=3mm]}},
            node/.style={circle,inner sep=1mm,minimum size=1cm,draw,black,very thick,text=black}]
            \node [node] (x) {$X$};
            \node [node] (y) [right of = x] {$Y$};
            \node [node] (h) [above of = x, xshift=8mm]  {$H$};
            \node [node] (a) [above of = x, xshift=-12mm, yshift=3mm]  {$A$};
            \path[normaledge] (a) edge (x);
            \path[normaledge] (a) edge (y);
            \path[normaledge] (a) edge (h);
            \path[normaledge] (x) edge (y);
            \path[normaledge] (h) edge (y);
            \path[normaledge] (x) edge (h);
        \end{tikzpicture}
    }
    \subfigure[]{
        \begin{tikzpicture}[auto, scale=1,
            node distance = 18mm and 6mm,
            normaledge/.style ={arrows=-{Latex[length=3mm]}},
            node/.style={circle,inner sep=1mm,minimum size=1cm,draw,black,very thick,text=black}]
            \node [node] (x) {$X$};
            \node [node] (y) [right of = x] {$Y$};
            \node [node] (h) [above of = x, xshift=8mm]  {$H$};
            \node [node] (a) [above of = x, xshift=-12mm, yshift=3mm]  {$A$};
            \path[normaledge] (a) edge (x);
            \path[normaledge] (a) edge (y);
            \path[normaledge] (a) edge (h);
            \path[normaledge] (y) edge (x);
            \path[normaledge] (h) edge (y);
            \path[normaledge] (x) edge (h);
        \end{tikzpicture}
    }
    
    \vskip 0.5cm
    
    \subfigure[]{
        \begin{tikzpicture}[auto, scale=1,
            node distance = 18mm and 6mm,
            normaledge/.style ={arrows=-{Latex[length=3mm]}},
            node/.style={circle,inner sep=1mm,minimum size=1cm,draw,black,very thick,text=black}]
            \node [node] (x) {$X$};
            \node [node] (y) [right of = x] {$Y$};
            \node [node] (h) [above of = x, xshift=8mm]  {$H$};
            \node [node] (a) [above of = x, xshift=-12mm, yshift=3mm]  {$A$};
            \path[normaledge] (a) edge (x);
            \path[normaledge] (a) edge (y);
            \path[normaledge] (a) edge (h);
            \path[normaledge] (h) edge (x);
            \path[normaledge] (y) edge (h);
            \path[normaledge] (x) edge (y);
        \end{tikzpicture}
    }
    \subfigure[]{
        \begin{tikzpicture}[auto, scale=1,
            node distance = 18mm and 6mm,
            normaledge/.style ={arrows=-{Latex[length=3mm]}},
            node/.style={circle,inner sep=1mm,minimum size=1cm,draw,black,very thick,text=black}]
            \node [node] (x) {$X$};
            \node [node] (y) [right of = x] {$Y$};
            \node [node] (h) [above of = x, xshift=8mm]  {$H$};
            \node [node] (a) [above of = x, xshift=-12mm, yshift=3mm]  {$A$};
            \path[normaledge] (a) edge (x);
            \path[normaledge] (a) edge (y);
            \path[normaledge] (a) edge (h);
            \path[normaledge] (y) edge (x);
            \path[normaledge] (y) edge (h);
            \path[normaledge] (h) edge (x);
        \end{tikzpicture}
    }

    \vskip 0.5cm
    
    \subfigure[]{
        \begin{tikzpicture}[auto, scale=1,
            node distance = 18mm and 6mm,
            normaledge/.style ={arrows=-{Latex[length=3mm]}},
            node/.style={circle,inner sep=1mm,minimum size=1cm,draw,black,very thick,text=black}]
            \node [node] (x) {$X$};
            \node [node] (y) [right of = x] {$Y$};
            \node [node] (h) [above of = x, xshift=8mm]  {$H$};
            \node [node] (a) [above of = x, xshift=-12mm, yshift=3mm]  {$A$};
            \path[normaledge] (a) edge (x);
            \path[normaledge] (a) edge (y);
            \path[normaledge] (a) edge (h);
            \path[normaledge] (h) edge (x);
            \path[normaledge] (h) edge (y);
            \path[normaledge] (x) edge (y);
        \end{tikzpicture}
    }
    \subfigure[]{
        \begin{tikzpicture}[auto, scale=1,
            node distance = 18mm and 6mm,
            normaledge/.style ={arrows=-{Latex[length=3mm]}},
            node/.style={circle,inner sep=1mm,minimum size=1cm,draw,black,very thick,text=black}]
            \node [node] (x) {$X$};
            \node [node] (y) [right of = x] {$Y$};
            \node [node] (h) [above of = x, xshift=8mm]  {$H$};
            \node [node] (a) [above of = x, xshift=-12mm, yshift=3mm]  {$A$};
            \path[normaledge] (a) edge (x);
            \path[normaledge] (a) edge (y);
            \path[normaledge] (a) edge (h);
            \path[normaledge] (y) edge (x);
            \path[normaledge] (h) edge (y);
            \path[normaledge] (h) edge (x);
        \end{tikzpicture}
    }
    
    \caption{All possible DAGs compatible with the anchor framework.}
    \label{fig:dag_variations}
\end{figure}

\section{Proofs}
\label{sec:proofs}

\subsection{Proof of Theorem~\ref{thm:ACloss}}
\begin{proof}
    Let's first note that from the SCM Eq.~ \ref{Eq:anchor-model} we have the following decomposition:
    \begin{align*}
        f(C_{X Y}) &= f(\mathbf{D}C_\varepsilon\mathbf{D}^T + \mathbf{D} \mathbf{M}C_A \mathbf{M}^T\mathbf{D}^T)\\
        &= f(\mathbf{D}C_\varepsilon\mathbf{D}^T) + f(\mathbf{D} \mathbf{M} C_A \mathbf{M}^T\mathbf{D}^T)
    \end{align*}
    by linearity of $f$. Thus when taking the supremum of the expectation of $f(C_{X Y})$ over $C^\gamma$, we get
    \begin{align*}
        \sup_{\nu \in C^\gamma} \Enu[f(C_{X Y})] = f(\mathbf{D}\Sigma_\varepsilon\mathbf{D}^T) + \sup_{\nu \in C^\gamma} f(\mathbf{D} \mathbf{M} \Sigma_\nu \mathbf{M}^T\mathbf{D}^T),
    \end{align*}
    since $f(\mathbf{D}\Sigma_\varepsilon\mathbf{D}^T)$ is not affected by the intervention. Here $\Enu[.]=\mathbb{E}_{(X, Y
    )\sim \Pnu }[.]$ the expectation for the intervention distribution $\Pnu$. Using the definition of $C^\gamma$ leads to 
    \begin{align}
    \begin{split}
\label{Eq:loss_decomp_final}
        \sup_{\nu \in C^\gamma} \Enu[f(C_{X Y})] &= f(\mathbf{D}\Sigma_\varepsilon\mathbf{D}^T) + \gamma f(\mathbf{D} \mathbf{M} \Sigma_A \mathbf{M}^T\mathbf{D}^T).
        \end{split}
    \end{align}
    Let's now note that for $(A, X, Y) \sim \mathbb{P}_{\text{train}}$ we have
    \begin{align*}
        \Etrain \left [\begin{pmatrix}X \\ Y\end{pmatrix} | A \right] &= \mathbf{D} \Etrain[\varepsilon|A] + \mathbf{D} \mathbf{M}\Etrain[A|A] \\
        &=  \mathbf{D}\mathbf{M}A,
    \end{align*}
    where $\mathbb{P}_{\text{train}}$ is the training distribution and $\Etrain[.] = E_{(X, Y)\sim \mathbb{P}_{\text{train}}}[.]$ the expectation with regard to $\mathbb{P}_{\text{train}}$.

    As $\varepsilon$ is mean centred and independent of $A$. Thus, we can write 
    \begin{align*}
        f(C_{X Y|A}) = f(\mathbf{D}\mathbf{M}C_A\mathbf{M}^T\mathbf{D}^T)
    \end{align*}
    by linearity of $f$. Taking its expectation over the training distribution, we get
    \begin{align}
\label{Eq:Decomp_right_term}
        \Etrain[f(C_{X Y| A }) ] = f(\mathbf{D}\mathbf{M}\Sigma_A\mathbf{M}^T\mathbf{D}^T),
    \end{align}
    which is similar to the right term of Eq.~\eqref{Eq:loss_decomp_final} as $\Etrain[f(C_{X Y| A }) ]= f(\Sigma_{X Y| A })$. A similar reasoning leads to 
    \begin{align}\label{Eq:Decomp_left_term}
      \Etrain[ f(C_{X Y})] -  \Etrain[f(C_{X Y|A})] &= f(\mathbf{D}\Sigma_\varepsilon\mathbf{D}^T).
    \end{align}
    
    Pluging-in Eq.~\eqref{Eq:Decomp_right_term} and Eq.~\eqref{Eq:Decomp_left_term} in Eq.~\eqref{Eq:loss_decomp_final} we get
    \begin{align*}
        \sup_{\nu \in C^\gamma} \Enu[f(C_{X Y})] &= \Etrain[ f(C_{X Y})]  + (\gamma-1) \Etrain [f(C_{X Y|A}) ]\\
       &= f(\Sigma_{X Y}) + (\gamma-1)  f(\Sigma_{X Y|A}),
    \end{align*}
    which concludes the proof.
\end{proof}

\subsection{Proof of Proposition }

\begin{proof}\label{pr:proof_proposition}
    The loss can easily decomposed as 
    \begin{align*}
        \mathbb{E}[f_{\mathbf{\Theta}}(R(\Bmax) \otimes R(\Bmax))] &= \mathbb{E}[f_{\mathbf{\Theta}}(R(\mathbf{B})\otimes R(\mathbf{B}))] + f_{\mathbf{\Theta}}((\mathbf{B} - \Bmax)\Sigma_{XYH}(\mathbf{B} - \Bmax)^T) \\
        &- \mathbb{E}[f_{\mathbf{\Theta}}(R(\mathbf{B}) \otimes (\mathbf{B} - \Bmax) \begin{pmatrix}X \\ Y \\ H\end{pmatrix})]\\
        &\geq \mathbb{E}[f_{\mathbf{\Theta}}(R(\mathbf{B})\otimes R(\mathbf{B}))] + f_{\mathbf{\Theta}}((\mathbf{B} - \Bmax)\Sigma_{XYH}(\mathbf{B} - \Bmax)^T) \\
        &- \mathbb{E}[f_{\mathbf{\Theta}}((R(\mathbf{B}) \otimes R(\mathbf{B})))]^{\frac{1}{2}}   \mathbb{E}[f_{\mathbf{\Theta}}(((\mathbf{B}- \Bmax)\Sigma_{XYH}(\mathbf{B} - \Bmax)^T))^{\frac{1}{2}}\\
    \end{align*}

    Where the first equality is a simple bias-variance decomposition and we use Cauchy-Schartz inequality for the inequality. By assumptions
    \begin{align*}
        \sup_{\nu \in C^{\gamma}} \Enu [f_{\mathbf{\Theta}}(R(\Bmax) \otimes R(\Bmax))] \leq \sup_{\nu \in C^{\gamma}} \Enu [f_{\mathbf{\Theta}}(R(\mathbf{B}) \otimes R(\mathbf{B}))].
    \end{align*}
    This leads to 
    \begin{align*}
        \Enu [f_{\mathbf{\Theta}}(R(\mathbf{B})\otimes R(\mathbf{B}))] &\geq \Enu [f_{\mathbf{\Theta}}(R(\mathbf{B})\otimes R(\mathbf{B}))] + f_{\mathbf{\Theta}}((\mathbf{B} - \Bmax)\Sigma_{XYH}^{do(A:=\nu)}(\mathbf{B} - \Bmax)^T) \\& - \Enu [f_{\mathbf{\Theta}}((R(\mathbf{B}) \otimes R(\mathbf{B}))) ]^{\frac{1}{2}} f_{\mathbf{\Theta}}(((\mathbf{B} - \Bmax)\Sigma_{XYH}^{do(A:=\nu)}(\mathbf{B} - \Bmax)^T))^{\frac{1}{2}}\\
    \end{align*}
    which is equivalent to 
    \begin{align*}
        f_{\mathbf{\Theta}}((\mathbf{B} - \Bmax)\Sigma_{XYH}^{do(A:=\nu)}(\mathbf{B} - \Bmax)^T) &\leq 4 \mathbb{E}[f_{\mathbf{\Theta}}(R(\mathbf{B}) \otimes R(\mathbf{B}))]\\
        &\leq 4f_{\mathbf{\Theta}}(\Sigma_\varepsilon) + 4\gamma f_{\mathbf{\Theta}}(\mathbf{M} \Sigma_A \mathbf{M}^T).
    \end{align*}
    This concludes the proof.
\end{proof}

\subsection{Proof of Equivalence for Transformed Data}

\begin{proof}
\label{pr:proof_estimator}
From 
\begin{align}\label{Eq:perturbed_data}
    \begin{split}
        \tilde{\mathbf{X}} &= (\mathbf{I} + (\sqrt{\gamma} - 1)\Pi_\mathbf{A}) \mathbf{X}\\
        \tilde{\mathbf{Y}} &= (\mathbf{I} + (\sqrt{\gamma} - 1)\Pi_\mathbf{A}) \mathbf{Y},
    \end{split}
\end{align}
 we can easily derive that 
 \begin{align*}
     \mathbf{S}_{\tilde{X}\tilde{Y}} = & \frac{1}{n-1} (\tilde{\mathbf{X}}\tilde{\mathbf{Y}})^T (\tilde{\mathbf{X}}\tilde{\mathbf{Y}})\\
     = &\frac{1}{n-1} ((\mathbf{X}\mathbf{Y} + (\sqrt{\gamma} - 1)\Pi_\mathbf{A}\mathbf{X}\mathbf{Y})^T (\mathbf{X}\mathbf{Y} + (\sqrt{\gamma} - 1)\Pi_\mathbf{A} \mathbf{X}\mathbf{Y}))\\
\end{align*}
As $\Pi_\mathbf{A}$ is an orthogonal projection, we have that $(\mathbf{X}\mathbf{Y})^T  (\Pi_\mathbf{A} \mathbf{X}\mathbf{Y}) = \Pi_\mathbf{A} (\mathbf{X}\mathbf{Y})^T\mathbf{X}\mathbf{Y} = \mathbf{S}_{XY|A}$, leading to
\begin{align*}
      \mathbf{S}_{\tilde{X}\tilde{Y}} = & \mathbf{S}_{XY} + 2(\sqrt{\gamma}-1)\mathbf{S}_{XY|A}
      + (\sqrt{\gamma}-1)^2\mathbf{S}_{XY|A}\\
      = &\mathbf{S}_{XY} + (\gamma-1)\mathbf{S}_{XY|A}.
 \end{align*}
Thus, by linearity of $f_{\mathbf{\Theta}}$, the minimiser of $f_{\mathbf{\Theta}}(\mathbf{S}_{\tilde{X}\tilde{Y}})$ is given by
 \begin{align*}
     \mathbf{\Theta}^\gamma &= \argmin_{\mathbf{\Theta}} f_{\mathbf{\Theta}}(\mathbf{S}_{\tilde{X}\tilde{Y}}) \\
    &= \argmin_{\mathbf{\Theta}} f_{\mathbf{\Theta}}(\mathbf{S}_{XY}) + (\gamma-1) f_{\mathbf{\Theta}}(\mathbf{S}_{XY|A}).
 \end{align*}
 This conclude the proof.
\end{proof}

\section{ESTIMATORS: FURTHER DETAILS}\label{sec:estimators_appendix}

\paragraph{Simpler Formulation}
For computational and practical reasons, $\widehat{\mathbf{\Theta}}^\gamma$ can be estimated by transforming the training data:
\begin{align}\label{Eq:perturbed_data_estimator}
    \tilde{\mathbf{X}} &= (\mathbf{I} + (\sqrt{\gamma} - 1)\Pi_\mathbf{A}) \mathbf{X} \\
    \tilde{\mathbf{Y}} &= (\mathbf{I} + (\sqrt{\gamma} - 1)\Pi_\mathbf{A}) \mathbf{Y},
\end{align}
where $\Pi_\mathbf{A} = \mathbf{A}(\mathbf{A}^T\mathbf{A})^{-1}\mathbf{A}^T \in \mathbb{R}^{n \times n}$ 
projects onto the column space of
$\mathbf{A}$, assuming that $\mathbf{A}^T\mathbf{A}$ is invertible and 
$\mathbf{X}$ and $\mathbf{Y}$ are centered. Finally, build the estimator $\widehat{\mathbf{\Theta}}^\gamma = \argmin_{\mathbf{\Theta}} f_{\mathbf{\Theta}}(\mathbf{S}_{\tilde{X}\tilde{Y}})$, where $\mathbf{S}_{\tilde{X}\tilde{Y}}$ is the empirical variance-covariance of the projected data defined in Eq. ~\eqref{Eq:perturbed_data_estimator}. Proof of the equivalence of using transformed data as in Eq.~\eqref{Eq:perturbed_data_estimator} is equivalent to Eq.~\eqref{Eq:estimator_full} is available in Supplementary ~\ref{pr:proof_estimator}. We give computational complexity and consistency results in Supplementary \ref{sec:estimators_appendix}.

\paragraph{Consistency}\label{sec:consistency}
From the law of large numbers, the empirical covariance matrix of $(X, Y, A)$ converges to its empirical covariance matrix, i.e.,
$\mathbf{S}_{X Y} \xrightarrow[n \to \infty]{} \Sigma_{XY}$.
Thus, the consistency of the anchor-regularised estimator depends on the consistency of the original estimator. For continuous functions $f$, we have by continuity that $\mathbf{\widehat{\Theta}}^\gamma \xrightarrow[n \to \infty]{} \mathbf{\Theta}^\gamma$ \citep[see ][section 4.1]{Rothenhäusler2018}.

\paragraph{Computational Complexity}\label{sec:compexity}
The computational cost of projecting $X$ and $Y$ into the span of $A$ involves computing the covariance matrix of $A$ (of cost $O(nr^2)$), inverting it (of cost $O(r^3)$), and two matrix products (of cost $O(nrp)$ and $O(nr^2)$), resulting in a total complexity of $O(nr^2 + nrp + r^3)$. Assuming the unregularised MVA algorithm's computational cost is $c$, the complexity of its anchor-regularised version is $O(c + nr^2 + nrp + r^3)$. We note that this generally incurs an affordable additive computational cost, as it is equivalent to two linear regressions. The only main concern arises with high-dimensional $A$, which is not addressed in this work.

\section{SIMULATION EXPERIMENTS: FURTHER DETAILS}
\subsection{Experiment Setting}\label{par:toy_experimental_setting}

The results of the simulation experiments are obtained as follows. The training data are sampled from the following SCM
\begin{align}
    \begin{split}
        \varepsilon_A, \varepsilon_H, &\varepsilon_X, \varepsilon_Y \sim \mathcal{N}(0, 1)\\
        A &\leftarrow \varepsilon_A\\
        H &\leftarrow \varepsilon_H\\
        X &\leftarrow A\mathbf{1}_p^\top  + H \mathbf{1}_p^\top + \varepsilon_X\\
        Y &\leftarrow \mathbf{W}^TX + H \mathbf{1}_d^\top + \varepsilon_Y,
    \end{split}
\end{align}
and the testing data are generated by modifying $\varepsilon_A$ such that $\varepsilon_A\sim \mathcal{N}(0, t)$. The perturbation strength $t$ is varied over a linear sequence $t\in [0, 4]$ with $20$ steps. 
We repeat each experiment $B=10$ times by sampling $n=300$ training and testing samples of $(A, X, Y)$.
We plot the average Mean Squared Errors in Fig.~\ref{fig:perturbation_strength} and Fig.~\ref{fig:perturbation_strength_non_gaussian_noise}. Both $A$ and $H$ are one-dimensional, while $X$ and $Y$ are $10$-dimensional. The matrix $\mathbf{W}$ is generated as a low-rank (of rank $\rho$) matrix such that $\mathbf{W} = \mathbf{U}\mathbf{V}$ with $\mathbf{A}\in \mathbb{R}^{d\times \rho}$ and $\mathbf{B}\in \mathbb{R}^{\rho\times p}$. 
The matrices $\mathbf{\Gamma}_i$ are hThe coefficients of $\mathbf{A}$ and $\mathbf{B}$ are sampled uniformly between $1$ and $2$ and are normalised such that their sum is $1$.

We employ the algorithms CCA, PLS, and MLR implemented in the \emph{scikit-learn} library (BSD 3-Clause License)~\cite{scikit-learn}, respectively \texttt{CCA}, \texttt{PLSRegression} and \texttt{LinearRegression} learners. We use the code available at \href{https://github.com/rockNroll87q/RRRR}{https://github.com/rockNroll87q/RRRR} (MIT License) ~\cite{SSB19, Mukherjee2011} for the Reduce Rank Regression algorithm and our implementation of OPLS based on \cite{Arenas-Garcia1015} using an eigenvalue decomposition as we use the constraint $A^TA=\mathbf{I}$.

\begin{figure}
    \centering
    \includegraphics[width=0.45\textwidth]{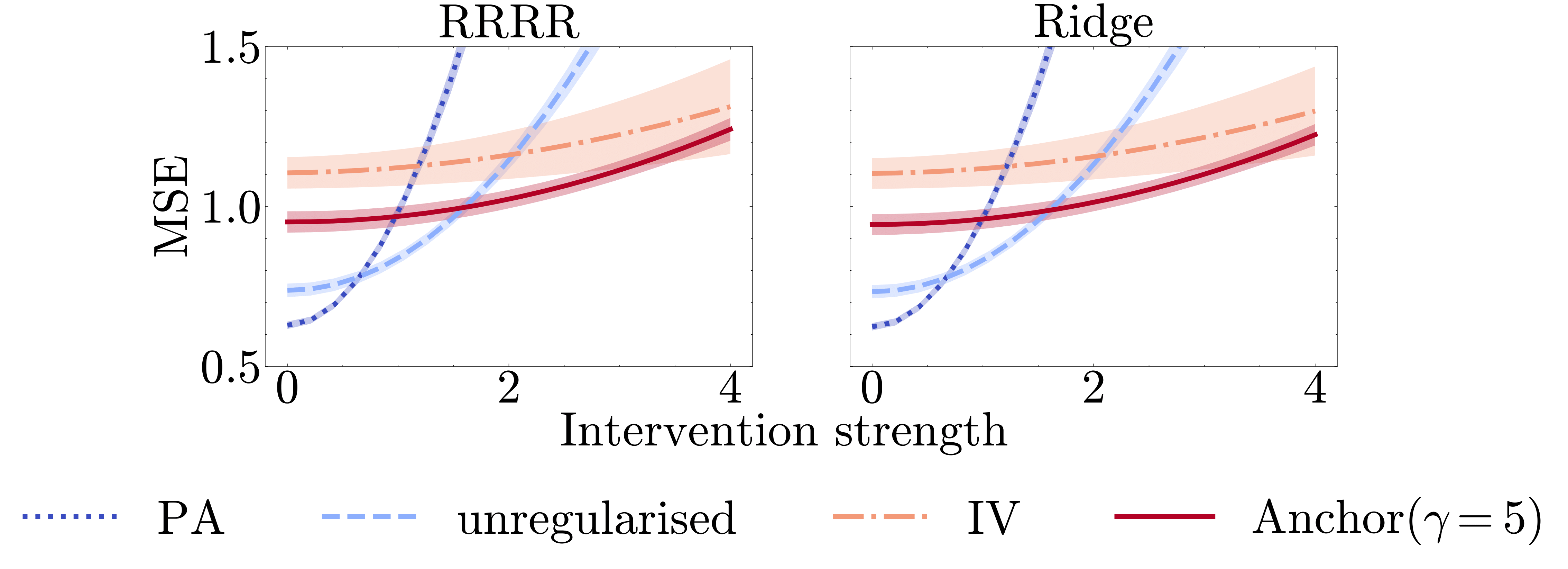}
    \caption{Experiments in high-dimensional setting with $d=p=300$ and $n=200$. We can see that both Multi-output Ridge Regression and Reduced rank Ridge Regression are optimal for a wide range of perturbation strength. Shaded areas represent the range of two standard errors of the mean for running $B=20$ times the experiment.}
    \label{fig:perturbation_strength_high_dim}
\end{figure}

\paragraph{High-Dimensional Setting} We also conduct a high-dimensional experiment to evaluate the performance of anchor-regularized algorithms when the dimensionality of $X$ and $Y$ exceeds the sample size. We generate data by sampling $n=200$ instances of $X$ and $Y$, each with a dimensionality of $d=p=300$. The rank of $\mathbf{W}$ is set to $\rho = 100$. We compare the results between Ridge Regression and Reduced Rank Ridge Regression. As shown in Figure \ref{fig:perturbation_strength_high_dim}, both methods demonstrate robustness across a wide range of perturbation strengths.

\paragraph{Non-Gaussian Noise Experiments}
To assess the robustness of our results presented in paragraph \ref{Par:Toy_model_results}, we conducted the same toy model experiments with noise $\varepsilon_A$, $\varepsilon_H$, $\varepsilon_X$, and $\varepsilon_Y$ following an exponential distribution (Fig.~\ref{fig:perturbation_strength_non_gaussian_noise}.A) and a gamma distribution (Fig.~\ref{fig:perturbation_strength_non_gaussian_noise}.B) with scales $1$.

As observed in Fig.~\ref{fig:perturbation_strength_non_gaussian_noise}, anchor-regularised models remain optimal for a wide range of perturbation strengths, except for anchor-regularised CCA. The behavior of anchor-regularised CCA is explained by its incompatibility with anchor regularisation.

\paragraph{Confounding Anchor Experiment}\label{sec:confounding_experiment}

We also reproduced experiments from equation 12 in \cite{Rothenhäusler2018} to demonstrate how anchor-regularised MVA algorithms exhibit interesting robustness properties when the anchor has a confounding effect. These results are illustrated in Fig.~\ref{fig:fig3_anchor_paper}. Here, the training distribution is encapsulated in the following Structural Causal Model (SCM):
\begin{align}
    \begin{split}
        \varepsilon_A,\, \varepsilon_H,\, &\varepsilon_X,\, \varepsilon_Y \sim \mathcal{N}(0, 1),\\
        A &\leftarrow \varepsilon_A,\\
        H &\leftarrow A + \varepsilon_H,\\
        X &\leftarrow A + H + \varepsilon_X,\\
        Y &\leftarrow \mathbf{W}^TX + H + \varepsilon_Y,
    \end{split}
\end{align}

with again testing distribution sampled by setting $\varepsilon_A \sim \mathcal{N}(0, t)$ where $t$ is the perturbation strength and $\mathbf{W}$ being lower rank.

\begin{figure}\label{fig:pareto-ARRRR}
    \centering
    \includegraphics[width=0.45\textwidth]{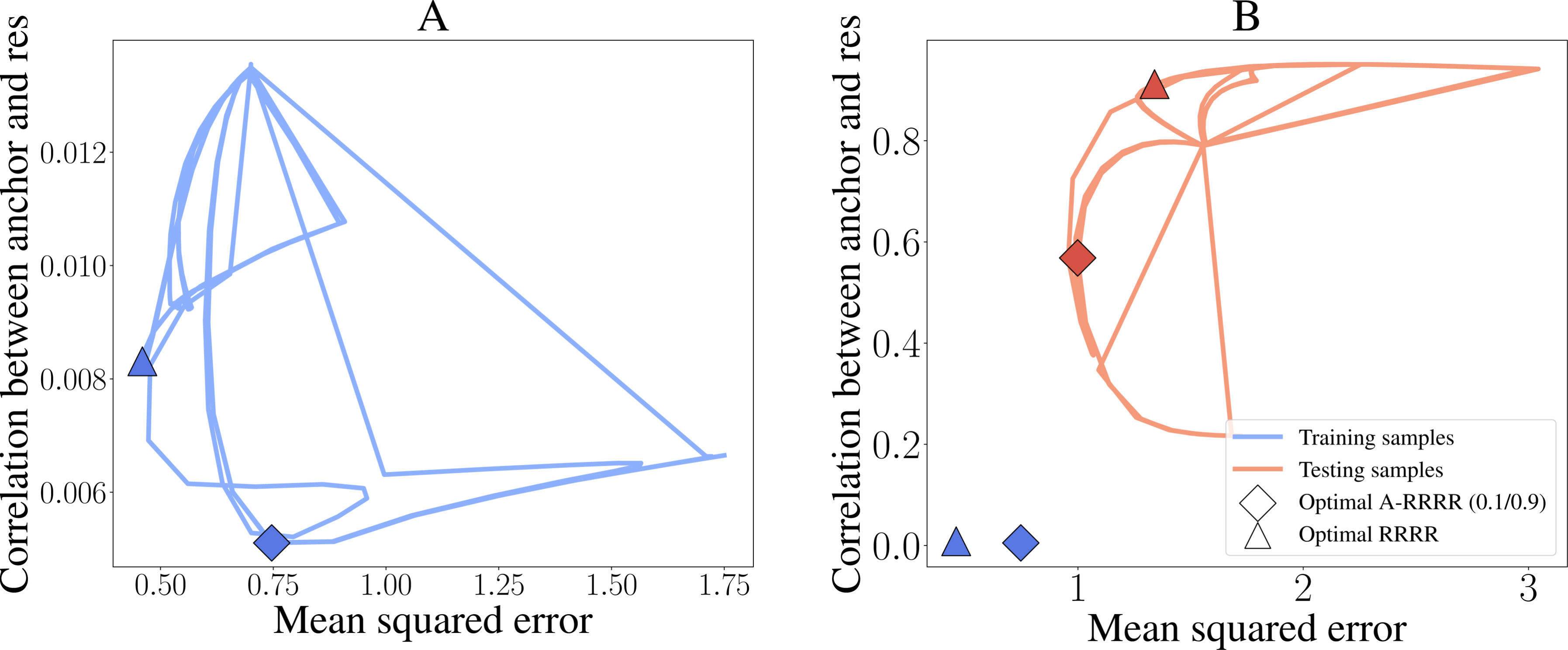}
    
    \caption{Pareto front for hyperparameters $(\gamma, \alpha, \rho)$ in RRRR and A-RRRR. For each pair $(\gamma, \alpha)$, $\rho$ is selected to minimise the weighted sum of objectives (equal weights given here).}
\end{figure}

We can see that all algorithms exhibit robustness properties for a very large range of perturbation strength. It is interesting to note that CCA presents a similar behavior as the \emph{anchor-compatible} algorithms. It would be interesting to investigate in which specific cases CCA gives robustness properties. 

\subsection{Hyperparameters Selection Experiment}\label{par:toy_high_dim_detail}

\begin{figure*}[t]
    \centering
    \includegraphics[width=0.9\textwidth]{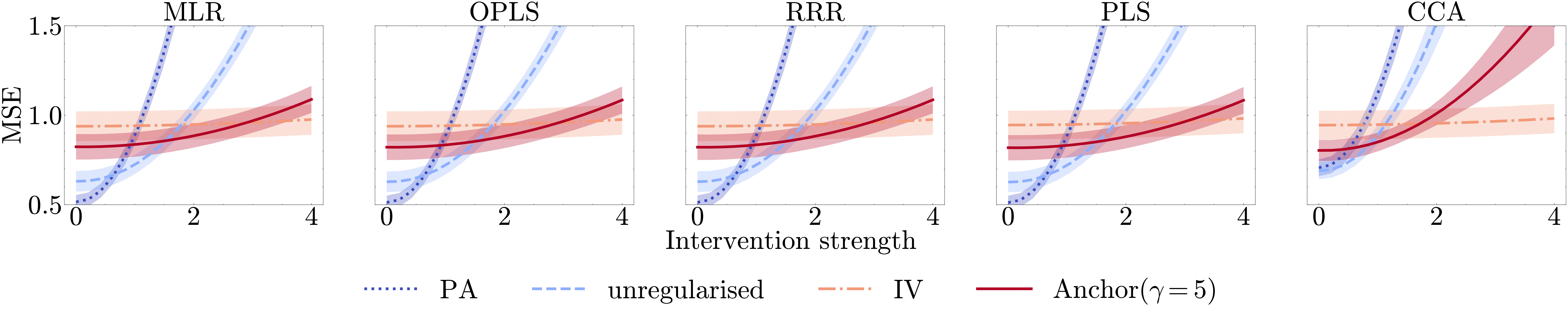}

    \includegraphics[width=0.9\textwidth]{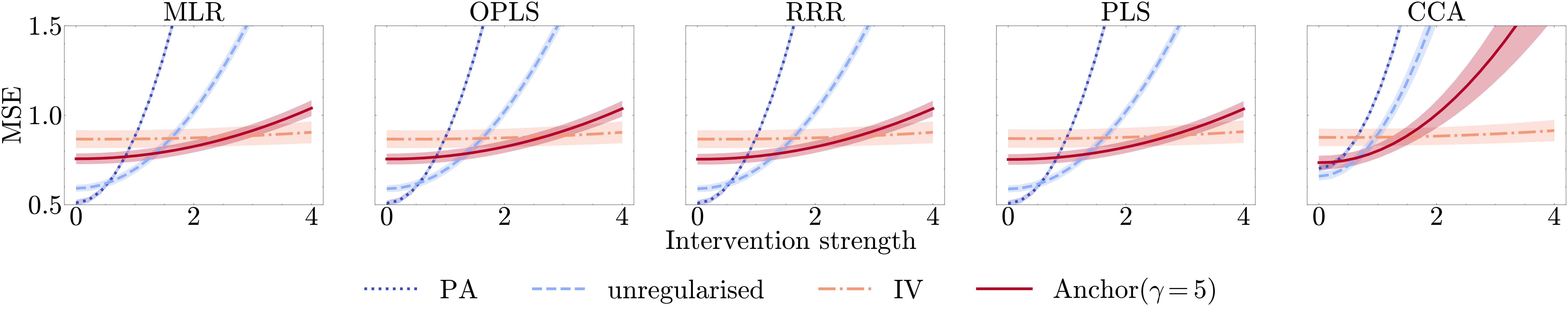}

    \caption{Similar experiments as in \ref{Par:Toy_model_results} with noise sampled from non-gaussian distributions. (Top) Exponential distribution of scale$1$, (Bottom) Poisson distribution of parameter $1$. We see that all the \emph{anchor-compatible} algorithms are robust to increasing perturbation strength. Shaded areas represent the range of two standard errors of the mean for running $B=20$ times the experiment.}
    \label{fig:perturbation_strength_non_gaussian_noise}
\end{figure*}

\begin{figure*}[t]
    \centering
    \includegraphics[width=0.9\textwidth]{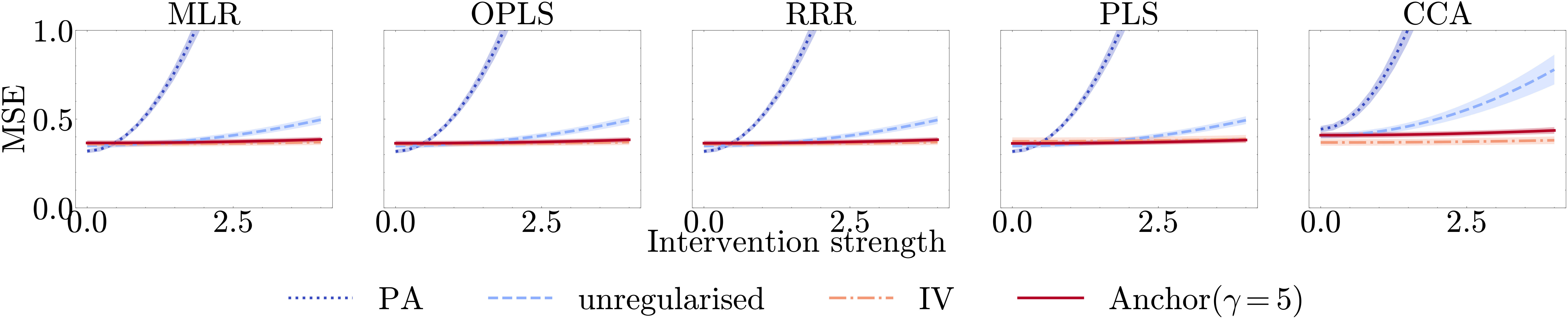}
    \caption{Experiment for anchor confounding $X$ and $Y$. We can see that all algorithms present a very robust behavior to increasing perturbation strength (even CCA which is not \emph{anchor-compatible}). Shaded areas represent the range of two standard errors of the mean for running $B=20$ times the experiment.}
    \label{fig:fig3_anchor_paper}
\end{figure*}

The results of the toy model experiments in the high-dimensional setting are obtained as follows. The training data are sampled following the DAG
\begin{align}
    \begin{split}
        \varepsilon_A,\, \varepsilon_H,\, &\varepsilon_X,\, \varepsilon_Y \sim \mathcal{N}(0, 1),\\
        A &\leftarrow \varepsilon_A,\\
        H &\leftarrow \varepsilon_H,\\
        X &\leftarrow \frac{1}{2}A + H + \varepsilon_X,\\
        Y &\leftarrow 2A + \mathbf{W}^TX + H + \varepsilon_Y,
    \end{split}
\end{align}

and the testing data are generated by modifying $\varepsilon_A$ such that $\varepsilon_A\sim \mathcal{N}(0, t)$. The perturbation strength $t$ is set to $2$. We repeat each experiment $B=10$ times by resampling $n=100$ training and validation samples and $n=400$ testing samples of $(A, X, Y)$. Both $A$ and $H$ are one-dimensional, while $X$ and $Y$ are $300$-dimensional. The matrix $\mathbf{W}$ is generated as a low-rank (of rank $\rho$ which is randomly sampled uniformly on $[10, 30]$) matrix such that $\mathbf{W} = \mathbf{U}\mathbf{V}$ with $\mathbf{U}\in \mathbb{R}^{d\times \rho}$ and $\mathbf{V}\in \mathbb{R}^{\rho\times p}$. The coefficients of $\mathbf{U}$ and $\mathbf{V}$ are sampled uniformly between $1$ and $3$ and are normalised such that their sum is $1$. Hyperparameters $\alpha$ ranging from $1$ to $10^5$ are chosen from a logarithmic scale with $20$ candidates, while $\rho$ ranges from $10$ to $30$ linearly with $10$ candidates, and $\gamma$ ranges from $1$ to $10^4$ logarithmically with $10$ candidates. Performance, measured in terms of Mean Squared Error and Mean correlation between anchor variable and residuals, is evaluated on a validation set (not seen during training) and a perturbed testing set. We illustrate the results by selecting the optimal rank for each pair $(\gamma, \alpha)$ (see Figure \ref{fig:pareto-ARRRR}). Optimal parameters are chosen to minimize different objectives: for RRRR, parameters $(\alpha, \rho)$ that minimize training MSE are selected, while for A-RRRR, a convex combination of correlation between anchor and residuals ($g_1(\gamma, \alpha, \rho)$) and MSE ($g_2(\gamma, \alpha, \rho)$) at training is minimized, i.e., $(\gamma, \alpha, \rho) = \argmin \sum_{i=1}^2 w_i \frac{g_i(\gamma, \alpha, \rho)}{\eta_i}$, where $\eta_1$ and $\eta_2$ rescale the two objectives. In Figure \ref{fig:pareto-ARRRR}, we present the results for optimal RRRR and A-RRRR with weights $w_1=0.1$ and $w_2=0.9$, emphasizing independence of anchor and residuals.

\begin{figure*}[!ht]
    \centering
    \includegraphics[width=0.9\textwidth]{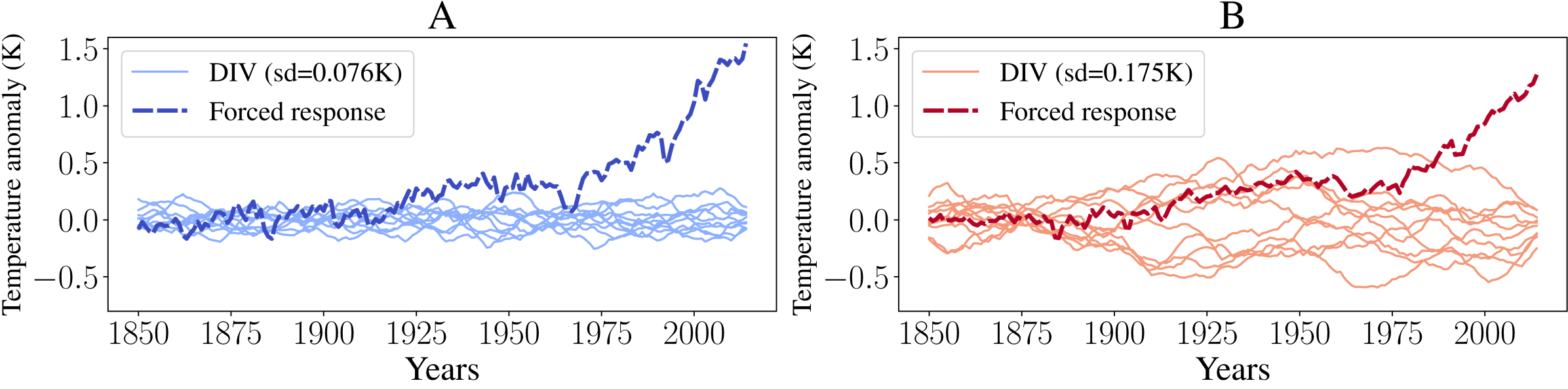}
    \caption{DIV and global forced response for a "low-variability" model (CESM2) (A) and for a "high-variability" model (CNRM-CM6-1) (B).}
    \label{fig:DIV_GMT_TRAIN_TEST}
\end{figure*}

\subsection{Anchor-MVA vs Univariate AR}\label{sec:AR_vs_AMVA}

As mentioned in the section, MVA methods offer a viable solution for handling correlated response variables within the context of the general multivariate linear model \(Y = \mu + \mathbf{W}X + \varepsilon\). When \(\mathbf{W}\) is not assumed to be full rank, the columns of \(Y\) exhibit interdependencies and often demonstrate discernible correlation patterns.

To clarify this point and the extent to which \emph{anchor-regularised} MVA is advantageous compared to simply running multiple AR, we conducted experiments using a similar setup as in \ref{Par:Toy_model_results}. We set the intervention strength \(t=2\), dimensions \(n=200, d=p=400\) with \(\mathbf{W}\) of rank \(\rho=10\), and used an anchor regularisation parameter \(\gamma=5\). We compared single variate AR (run for each output) and \emph{anchor-regularised} Reduced Rank Regression with an oracle for the rank. The experiment was run \(B=50\) times, and we obtained the following results in terms of MSE:

\begin{table}[h]
    \centering
    \caption{Comparaison of multiple AR and A-RRR with $\gamma=5$. Results are reported in term of MSE with 95\% confidence intervals.}
    \begin{tabular}{lcc}
        \toprule
         & Multiple AR (\(\gamma=5\)) &  A-RRR (\(\gamma=5\)) \\
        \midrule
         \textbf{Mean Squared Error} & \(1.91 \pm 0.26\) & \(\mathbf{1.45 \pm 0.24}\) \\
        \bottomrule
    \end{tabular}
    \label{tab:ARvsAMVA}
\end{table}

This showcases how the use of the \emph{anchor-regularised} algorithm can outperform AR when faced with high-dimensional output variables.

\section{Details of Real-World Experiment}\label{sec:real_world_details}

\subsection{Robust Climate Prediction}

\paragraph{Optimal Fingerprint for Detection of Forced Warming}

In D\&A studies, the optimal fingerprinting process involves several steps. Firstly, the response of the climate system to an external forcing is extracted using a statistical learning model to predict the forced climate response, denoted as $Y^{\text{forced}}_{\text{mod}}$. This is achieved by utilising spatial predictors from a gridded field of climate variables $X_{\text{mod}}$, following the regression equation:
\begin{align}
    Y_{\text{mod}}^{\text{forced}} = X_{\text{mod}}\beta + \beta_0 + \varepsilon.
\end{align}
Here, the spatial fingerprint is represented by the regression coefficient $(\beta, \beta_0)$.
Subsequently, a detection metric is obtained by projecting observations ($X_{\text{obs}}$) and unforced simulations ($X_{\text{cntl}}$) (known as control scenarios) onto the extracted fingerprint $\hat{\beta}$. Practically, as multiple members and models are available for the unforced scenario, a distribution for $X_{\text{mod}}^{\text{cntl}} \hat{\beta} + \hat{\beta}_0$ can be obtained. Then, it is tested whether $X_{\text{obs}} \hat{\beta} + \hat{\beta}_0$ lies within the same distribution. If the test is rejected, a forced response is detected.

\paragraph{From Global to Regional}
Moving from a global to regional scale presents challenges in D\&A studies. One of the current challenges is transitioning from detecting a global forced response to a regional or local forced response. This is because climate variability magnitude is larger at regional scales, and even larger at a local scale, leading to a lower signal-to-noise ratio of the forced warming response. As the signal-to-noise ratio is in this case much lower, it becomes much more challenging to detect the forced signal. Additionally, different climate models and observations exhibit different patterns of internal variability at regional scales, necessitating the training of statistical learning models robust to potential distribution shifts in internal variability.

\paragraph{Robustness to Multidecadal Internal Variability}

\begin{figure}[t]
    \centering
    \begin{tikzpicture}[auto, scale=12,
    node distance = 18mm and 6mm,
    edge/.style ={arrows=-{Latex[length=3mm]}},
    dashededge/.style ={dashed}, 
    node/.style={circle,inner sep=1mm,minimum size=1cm,draw,black,very thick,text=black}]

     \node [node] (x) {$X$};
     \node [node] (y) [right of = x, xshift=12mm] {$Y^{\text{forced}}_{\text{mod}}$};
     \node [node] (a) [left of = x, xshift=-12mm]  {$DIV$};

     \path[edge] (a) edge (x);


     \path[edge] (y) edge (x);
     \path[dashededge, bend right=20] (x) edge node[below] {$\mathbf{W}$} (y);

     
    \end{tikzpicture}
    \caption{DAG considered for detection of forced warming response as proposed in \cite{Sippel2021}.}
    \label{fig:dag_DIV_Y_X}
\end{figure}
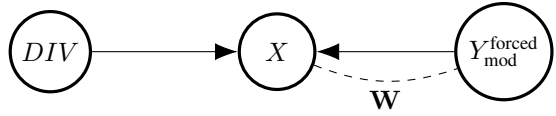

The ability to detect forced warming in Detection and Attribution studies is highly influenced by the level of internal variability \citep[see ][]{Parsons2020}. For instance, the observed 40-year Global Mean Temperature (GMT) trend of 0.76°C for the period $1980-2019$ would exceed the standard deviation of natural internal variability in CMIP5 and CMIP6 archive models classified as "low-variability" by a factor of $5$ or more \citep[see Fig. 1B in][]{Sippel2021}. However, for "high-variability" models, this trend exceeds the standard deviation only by a factor of $2$. This becomes evident when comparing models with "low-variability" (Figure \ref{fig:DIV_GMT_TRAIN_TEST} A) to those with "high-variability" (Figure \ref{fig:DIV_GMT_TRAIN_TEST} B), where it is less clear that internal variability alone could generate the Global Mean Temperature. This discrepancy poses a significant challenge, as in the former case, the observed warming has an extremely low probability of being generated by internal variability alone. Conversely, in the latter case, the rejection of the hypothesis that current temperature trends are solely attributable to internal variability is less robust.
For this reason, it is important to develop tests that are robust to changes in the magnitude of internal variability, particularly with the increasing magnitude of DIV. Assuming the DAG depicted in Figure \ref{fig:dag_DIV_Y_X}, anchor regularisation emerges as a suitable approach, enabling robust estimation robust to DIV increase.

\paragraph{Climate Models}
In this experiment, we utilised $7$ climate models from the CMIP5 and CMIP6 archives, each having at least $8$ members. The selection of these models was primarily practical, based on our access to them and their suitability for the train/test split procedure based on DIV magnitude.

\begin{table*}[t]
    \centering
    \renewcommand{\arraystretch}{0.9} 
    \small
    \caption{Climate models used in this work, their number of members, associated DIV magnitude and corresponding set (train or test).}
    \begin{tabular}{lcccc}
        \toprule
        Model & CMIP   &  $\sharp$members & DIV mag. & Set \\
        \midrule
         {CCSM4} & 5   & 8 & 0.052 & Train \\
         {NorCMP1} & 6   &  30 & 0.064 & Train \\
         {CESM2} & 6   &  11 & 0.076 & Train \\
         {HadCM3} & 5   &  10 & 0.072 & Train \\
         {CNRM-CM6-1} & 6   &  30 & 0.175 & Test \\
         {CNRM-ESM2-1} & 6   &  10 & 0.174 & Test \\
         {IPSL-CM6A-LR} & 6   &  32 & 0.141 & Test \\
        \bottomrule
    \end{tabular}
    \label{tab:CMIP56_models}
\end{table*}

\paragraph{Results}

\begin{table*}[t]
    \centering
    \caption{Metrics for RRRR, and they anchor-regularised version with $\gamma=5$ and $\gamma=100$. We can see that both anchor-regularised approaches outperformed unregularised RRRR in term of $R^2$ score and mean correlation of residuals with DIV.}
    
    \begin{tabular}{lcccc}
        \toprule
         & \multicolumn{2}{c}{Test}    & \multicolumn{2}{c}{Train}  \\
         \cmidrule{2-3}\cmidrule{4-5}
          & $R^2$    & Mean correlation &   $R^2$    & Mean correlation  \\
        \midrule
         RRRR & 0.506 &  0.419 &  \textbf{0.510} &  0.157\\
        A-RRRR ($\gamma=5)$ & \textbf{0.537} &  0.248 & 0.500 &  0.120 \\
        A-RRRR ($\gamma=100)$& 0.533 &  \textbf{0.123} & 0.487 &  \textbf{0.096}\\
        \bottomrule
    \end{tabular}
    \label{tab:metrics_rrrr}
\end{table*}

\begin{figure}[!t]
    \centering
    \includegraphics[width=0.40\textwidth]{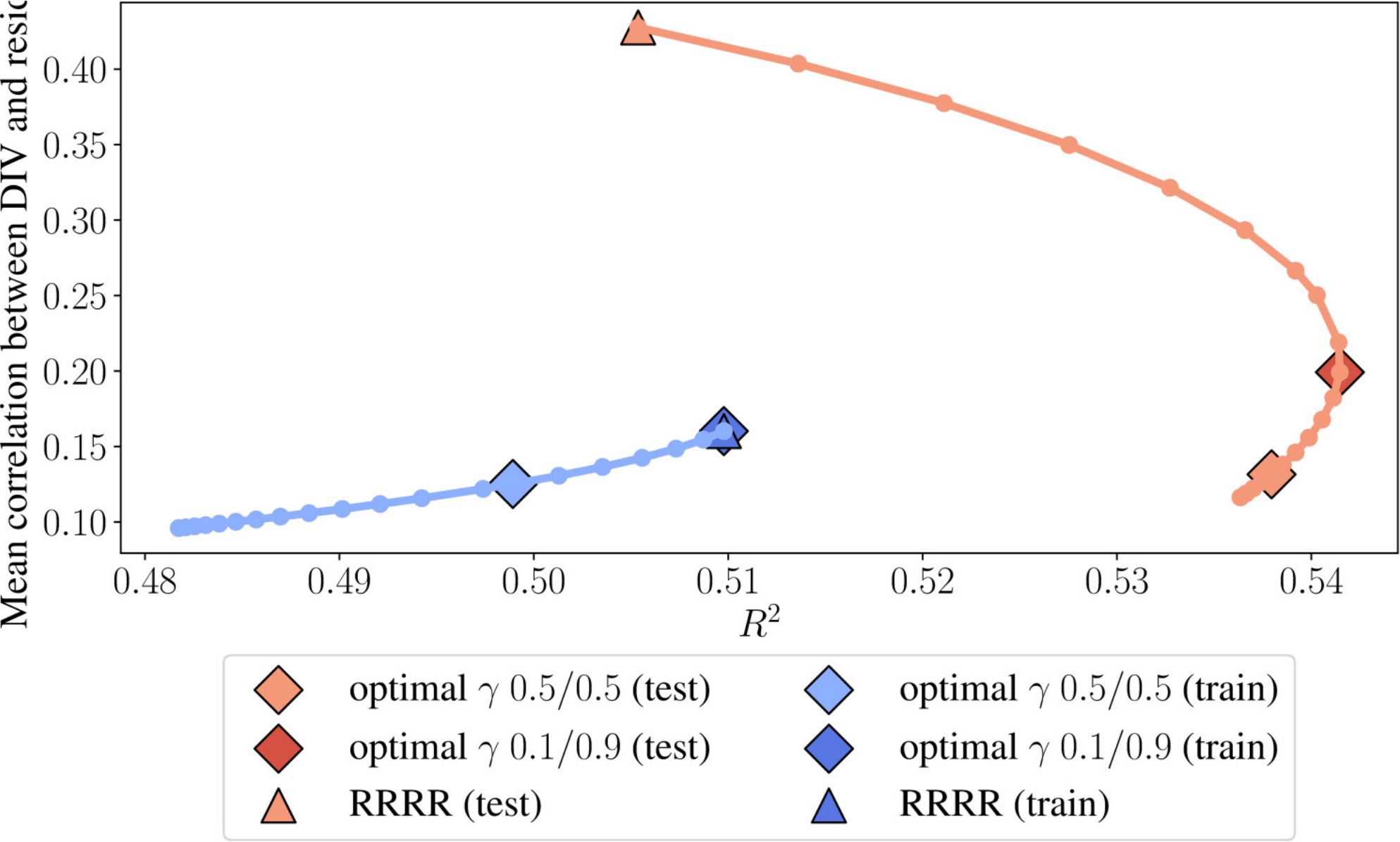}
    \caption{Pareto front for A-RRRR with optimal $\alpha$, $\rho$, and $\gamma \in [1, 10^2]$ with optimal $\gamma$ of $0.5/0.5$ $0.1/0.9$ weighted objectives.}
    \label{fig:pareto_gamma_opt_CMIP56}
\end{figure}

\begin{figure*}
    \centering
    \includegraphics[width=0.45\textwidth]{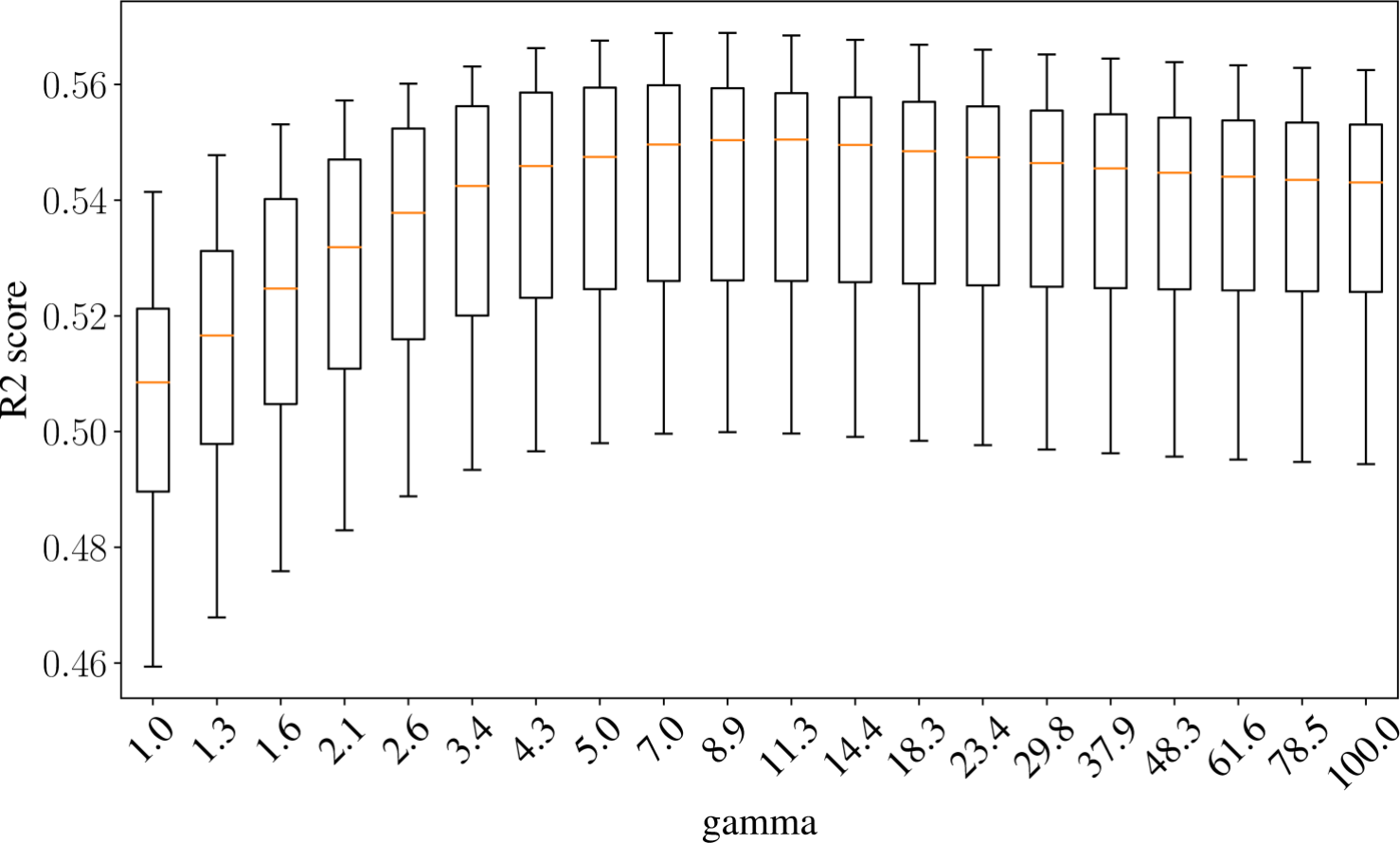}
    \includegraphics[width=0.45\textwidth]{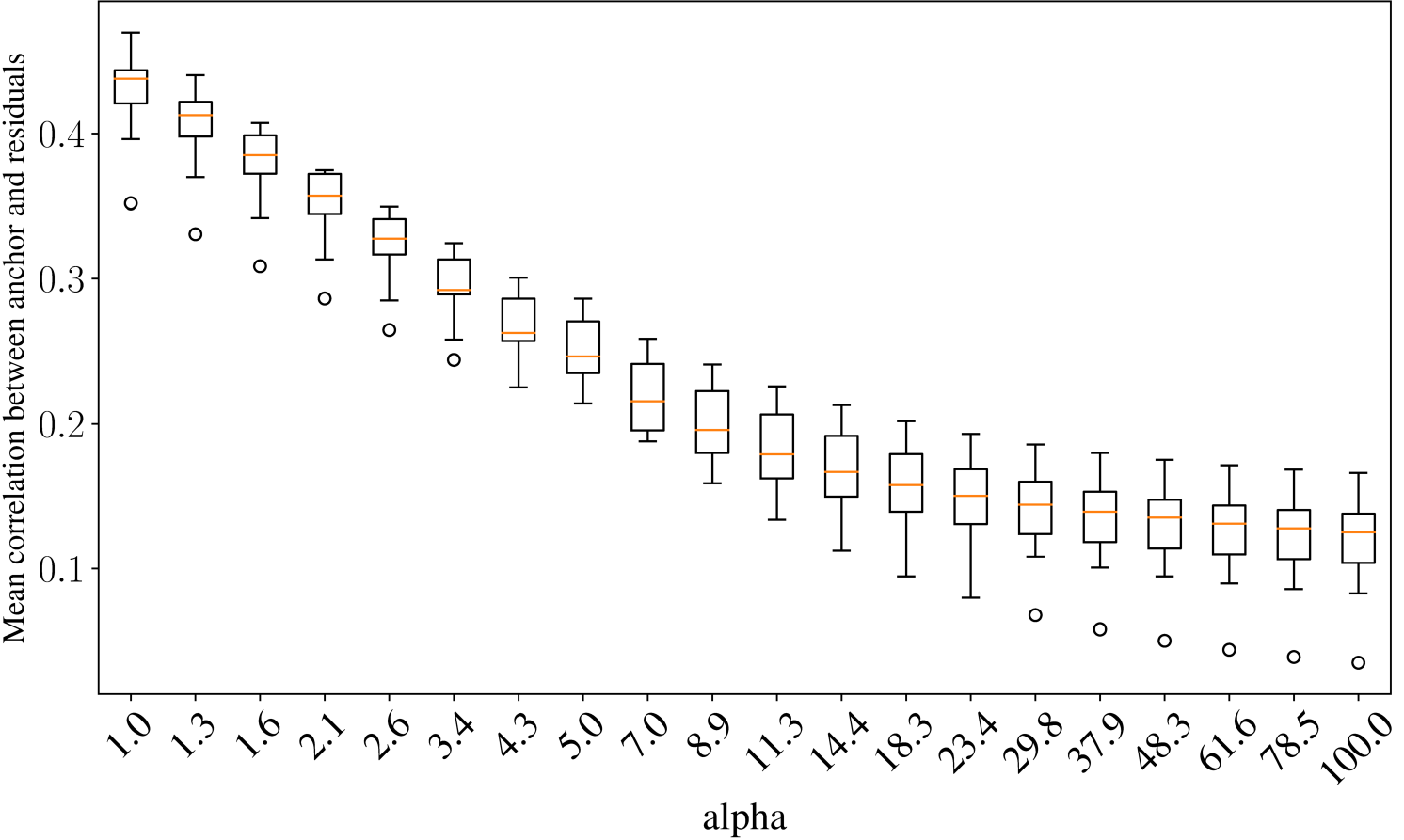}
    \caption{(Left) $R^2$ scores (averaged uniformly). (Right) Mean correlation between DIV and residuals for different values of $\gamma$. }
    \label{fig:corr_alpha}
\end{figure*}

The anchor-regularised version of RRRR generally exhibits superior performance compared to its unregularised counterpart (refer to Table~\ref{tab:metrics_rrrr} and Figure~\ref{fig:corr_alpha}), both in terms of prediction performance (measured via $R^2$) and correlation between residuals and DIV at testing time. Notably, these results are achieved at a lower training time cost in terms of $R^2$, particularly when utilising $\gamma = 5$. This trade-off between performance during training and testing becomes clear when looking at the pareto front emerging (Figure ~\ref{fig:pareto_gamma_opt_CMIP56}) when considering different $\gamma$ values for optimal hyperparameters $(\alpha, \rho)$ selected via the validation procedure detailed earlier. We can see that selecting $\gamma$ such as it minimize an equally weighted ($\gamma 0.5/0.5)$ objectives (prediction performances and correlation between residuals and DIV) lead to a very small decrease during training but induce strong robustness (being close to optimal) at testing in terms of predicting performance.

Regarding the spatial patterns of prediction performance, anchor-regularised RRRR outperforms the unregularised version, particularly in the northern hemisphere and notably at the North Pole, where internal climate variability is significantly higher than closer to the equator. It also exhibits superior performance in northern regions where unregularised RRRR performs poorly, albeit with suboptimal results. Conversely, it demonstrates slightly better performance in the southern hemisphere, especially in regions where the DIV and the residuals exhibit anticorrelation (e.g., central Africa and the southern region of South America). In terms of the correlation between DIV and residuals, anchor-regularised RRRR outperforms the unregularised version across most regions, except for central Africa and the southern regions of South America, where residuals and DIV exhibit anticorrelation. On average, it reduces the correlation between residuals and DIV by more than $15\%$ for $\gamma=5$ and close to $30\%$ for $\gamma=100$.


\begin{figure*}
    \centering
    \includegraphics[width=0.8\textwidth]{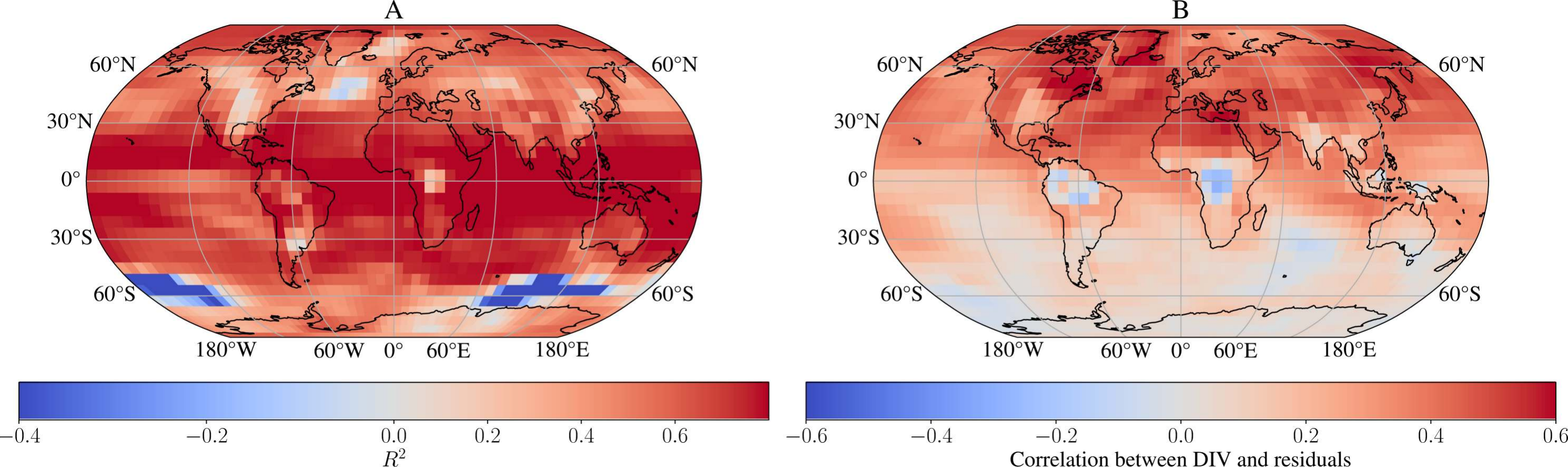}
    \label{fig:R2_corr_plot_gamma5}
    \caption{A-RRRR ($\gamma=5$) scores. (A)  $R^2$ scores. (B) Correlation between DIV and residuals.}
\end{figure*}

\begin{figure*}
    \centering
    \includegraphics[width=0.8\textwidth]{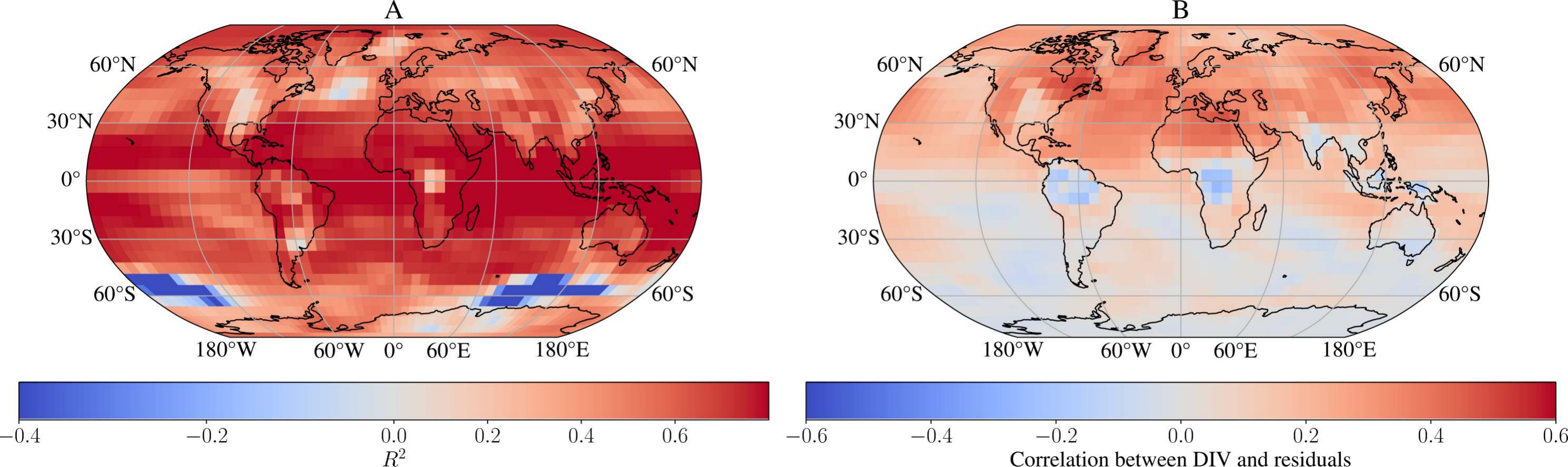}
    \caption{A-RRRR ($\gamma=100$) scores. (A)  $R^2$ scores. (B) Correlation between DIV and residuals.}
    \label{fig:R2_corr_plot_gamma100}
\end{figure*}

\begin{figure*}
    \centering
    \includegraphics[width=0.8\textwidth]{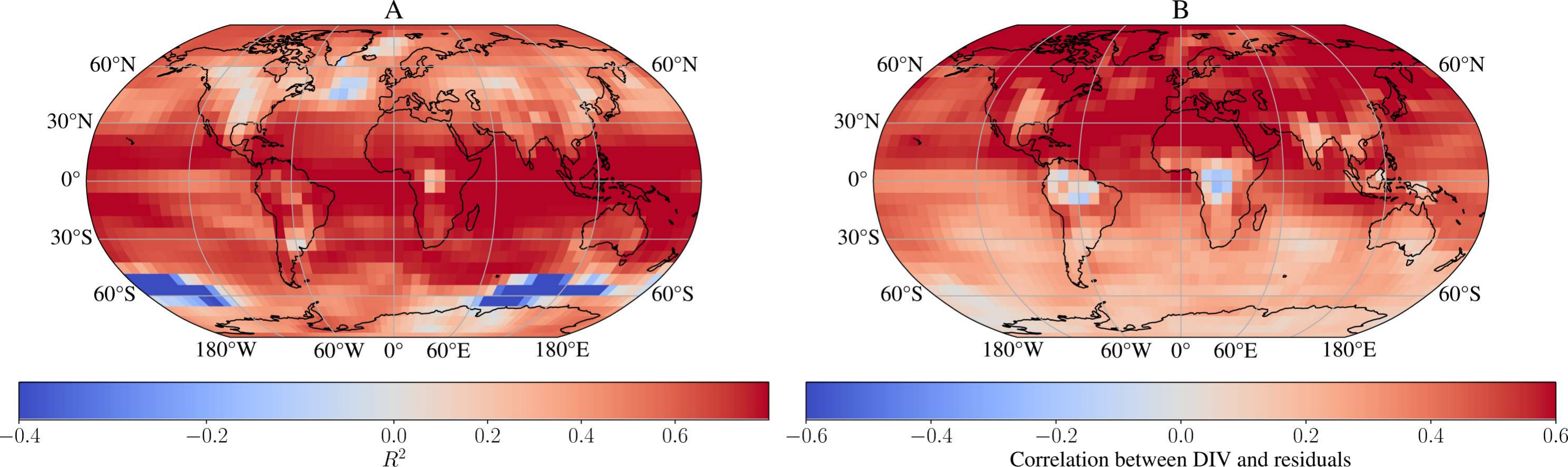}
    \caption{Unregularised RRRR scores. (A)  $R^2$ scores. (B) Correlation between DIV and residuals.}
    \label{fig:R2_corr_plot_rrrr}
\end{figure*}


\newpage

\subsection{Robust air-quality prediction}

\paragraph{IRM and CVP training} IRM and CVP models are trained to MSE convergence with a learning rate of $0.1$, patience of 200 epochs, tolerance of $10^{-4}$, and a maximum of 50,000 epochs.

We provide pseudocodes for both CVP (Alg. \ref{alg:CVP}) and IRM (Alg. \ref{alg:IRM}) algorithms.

\begin{algorithm}[H]
\label{alg:IRM}
\caption{Invariant Risk Minimization Linear Regression}
\begin{algorithmic}[1]
    \State \textbf{Initialize:} Parameters $\phi$, $w$, learning rate $\eta$
    \State \textbf{Input:} Training data $X$, $Y$, environment labels $E$, regularization $\lambda_{irm}$, epochs $N$, patience $p$
    \State Set best loss to $\infty$, patience counter to 0
    \For{$epoch = 1$ to $N$}
        \State Compute IRM loss:
        \State \quad $\mathcal{L} = 0, P = 0$
        \For{each environment $e$ in $E$}
            \State Extract $X_e, Y_e$ where $E = e$
            \State Transform input: $X_e' = X_e \phi$
            \State Compute predictions: $\hat{Y}_e = X_e' w$
            \State Compute MSE loss: $\ell_e = \text{MSE}(\hat{Y}_e, Y_e)$
            \State Compute gradient penalty: $P += \left( \frac{d \ell_e}{d w} \right)^2$
            \State Update total loss: $\mathcal{L} += \ell_e$
        \EndFor
        \State Compute final loss: $\mathcal{L} = \mathcal{L} + \lambda_{irm} P$
        \State Update $\phi$ and $w$ using gradient descent
        \State Compute validation MSE for early stopping
        \If{validation MSE improves}
            \State Store best $\phi$, $w$
            \State Reset patience counter
        \Else
            \State Increment patience counter
        \EndIf
        \If{patience counter $\geq p$}
            \State Stop training
        \EndIf
    \EndFor
    \State Return best $\phi$ and $w$
\end{algorithmic}
\end{algorithm}

\begin{algorithm}
\label{alg:CVP}
\caption{Conditional Variance Penalty Linear Regression}
\begin{algorithmic}[1]
    \State \textbf{Initialize:} Model parameters $w$, learning rate $\eta$
    \State \textbf{Input:} Training data $X$, $Y$, environment labels $E$, regularization $\lambda_{cvp}$, epochs $N$, patience $p$
    \State Set best loss to $\infty$, patience counter to 0
    \For{$epoch = 1$ to $N$}
        \State Compute CVP loss:
        \State \quad $\mathcal{L} = \text{MSE}(Xw, Y)$
        \For{each environment $e$ in $E$}
            \State Extract $X_e, Y_e$ where $E = e$
            \State Compute predictions: $\hat{Y}_e = X_e w$
            \State Compute variance of predictions: $V_e = \text{Var}(\hat{Y}_e)$
            \State Update total loss: $\mathcal{L} += \lambda_{cvp} \sum V_e$
        \EndFor
        \State Update $w$ using gradient descent
        \State Compute validation MSE for early stopping
        \If{validation MSE improves}
            \State Store best $w$
            \State Reset patience counter
        \Else
            \State Increment patience counter
        \EndIf
        \If{patience counter $\geq p$}
            \State Stop training
        \EndIf
    \EndFor
    \State Return best $w$
\end{algorithmic}
\end{algorithm}



\begin{table*}
    \centering
    \begin{tabular}{llp{6cm}}
        \toprule
        Feature Name & Data Type & Description \\
        \midrule
        CO(GT) & Integer & True hourly averaged concentration CO in mg/m\textsuperscript{3}  \\
        PT08.S1(CO) & Categorical & Hourly averaged sensor response  \\
        NMHC(GT) & Integer & True hourly averaged overall Non-Metanic HydroCarbons concentration in \textmu g/m\textsuperscript{3} \\
        C6H6(GT) & Continuous & True hourly averaged Benzene concentration in \textmu g/m\textsuperscript{3}  \\
        PT08.S2(NMHC) & Categorical & Hourly averaged sensor response   \\
        NOx(GT) & Integer & True hourly averaged NOx concentration in ppb  \\
        PT08.S3(NOx) & Categorical & Hourly averaged sensor response   \\
        NO2(GT) & Integer & True hourly averaged NO2 concentration in \textmu g/m\textsuperscript{3} \\
        \bottomrule
    \end{tabular}
    \caption{Air Quality Dataset Outcome Description}
    \label{tab:vairable_pollution}
\end{table*}